\documentclass[letterpaper]{article} % DO NOT CHANGE THIS
\usepackage{aaai24}  % DO NOT CHANGE THIS
\usepackage{times}  % DO NOT CHANGE THIS
\usepackage{helvet}  % DO NOT CHANGE THIS
\usepackage{courier}  % DO NOT CHANGE THIS
\usepackage[hyphens]{url}  % DO NOT CHANGE THIS
\usepackage{graphicx} % DO NOT CHANGE THIS
\usepackage{natbib}  % DO NOT CHANGE THIS AND DO NOT ADD ANY OPTIONS TO IT
\usepackage{caption} % DO NOT CHANGE THIS AND DO NOT ADD ANY OPTIONS TO IT
\usepackage{float}

\usepackage[table,xcdraw]{xcolor}
\usepackage{amssymb,amsmath}
\usepackage{adjustbox}
\usepackage{multirow}
\usepackage{arydshln}
\usepackage{algorithmicx}
\usepackage[noend]{algpseudocode}
\usepackage{caption}
\usepackage{cuted}
\usepackage{algorithm}
\usepackage{newfloat}
\usepackage{listings}
\DeclareCaptionStyle{ruled}{labelfont=normalfont,labelsep=colon,strut=off} % DO NOT CHANGE THIS
\floatstyle{ruled}
\newfloat{listing}{tb}{lst}{}
\floatname{listing}{Listing}
\title{Federated Modality-specific Encoders and Multimodal Anchors\\ for Personalized Brain Tumor Segmentation}
\author{
    Qian Dai\textsuperscript{\rm 1,\equalcontrib}, %\thanks{With help from the AAAI Publications Committee.}\\
    Dong Wei\textsuperscript{\rm 2,\equalcontrib},
    Hong Liu\textsuperscript{\rm 2,3,\equalcontrib},
    Jinghan Sun\textsuperscript{\rm 2,3},
    Liansheng Wang\textsuperscript{\rm 1,\thanks{Corresponding author.}},
    Yefeng Zheng\textsuperscript{\rm 2}
}
\affiliations{
    \textsuperscript{\rm 1}School of informatics, Xiamen University, Xiamen, China\\%\hspace{8mm}
    \textsuperscript{\rm 2}Jarvis Research Center, Tencent Youtu Lab / Tencent Healthcare (Shenzhen) Co., Ltd., Shenzhen, China\\
    \textsuperscript{\rm 3}School of Medicine, Xiamen University, Xiamen, China\\  
    {\{daiqian,liuhong,jhsun\}@stu.xmu.edu.cn}, 
    {lswang@xmu.edu.cn},
    {\{donwei,yefengzheng\}@tencent.com},
}

\usepackage{bibentry}
\usepackage[table,xcdraw]{xcolor}
\usepackage{amssymb,amsmath}
\usepackage{adjustbox}
\usepackage{multirow}
\usepackage{arydshln}
\usepackage{algorithmicx}
\usepackage[noend]{algpseudocode}
\colorlet{red}{black}
\colorlet{blue}{black}
\begin{document}

\maketitle

\begin{abstract}
%Federated learning (FL) is particularly suitable for privacy-sensitive applications such as medical image analysis (MIA), by enabling the participant health units to collaboratively train a global model on their collective data without breaching privacy.
%%
Most existing federated learning (FL) methods for medical image analysis only considered intramodal heterogeneity, limiting their applicability to multimodal imaging applications.
In practice, it is not uncommon that some FL participants only possess a subset of the complete imaging modalities, 
% e.g., due to varying institutional protocols, 
posing inter-modal heterogeneity as a challenge to effectively training a global model on all participants' data.
In addition, each participant would expect to obtain a personalized model tailored for its local data characteristics from the FL in such a scenario.
% However, both demands cannot be solved by the conventional FedAvg pipeline.
%Fully aligned and paired multimodal image data have been shown to be effective in brain tumor segmentation. However, it is impractical for some clinical institutions to collect fully aligned paired complete modal data, and the problem of missing modality images can cause significant degradation in segmentation performance. 
%%
%The emergence of federated learning (FL) provides us with a new idea, allowing multiple clinical institutions to train a global model collaboratively without sharing data. And its good privacy security has been well applied in the biomedical field.
%%
In this work, we propose a new FL framework with federated modality-specific encoders and multimodal anchors (FedMEMA) to simultaneously address the two concurrent issues.
Above all, FedMEMA employs an exclusive encoder for each modality to account for the inter-modal heterogeneity in the first place.
In the meantime, while the encoders are shared by the participants, the decoders are personalized to meet individual needs.
Specifically, a server with full-modal data employs a fusion decoder to aggregate and fuse representations from all modality-specific encoders, thus bridging the modalities to optimize the encoders via backpropagation reversely.
Meanwhile, multiple anchors are extracted from the fused multimodal representations and distributed to the clients in addition to the encoder parameters.
On the other end, the clients with incomplete modalities calibrate their missing-modal representations toward the global full-modal anchors via scaled dot-product cross-attention, making up the information loss due to absent modalities while adapting the representations of present ones.
%In this paper, we discuss a new problem of multimodal federated learning. Although multimodal data always benefits from the complementarity of different modalities, different clinical institutions have data of different modes due to the modality difference.
%%(discrepancy)
%This modality difference further leads to modality heterogeneity, which is difficult to solve with the existing FL methods.
%%
%To solve this problem, we propose a personalized multimodal federated framework that utilizes a small amount of paired multimodal data on the server to help each uni-modal client perform Local Calibration via Cross-Attention, denoted as mFLCA.
%%
%Concretely, we personalize the parameters of the decoder and the server obtains the modal-specific encoder.
%Considering that the local client only has a single modal data, it has alternative attention on the global features, we use the cross-attention mechanism for local calibration, which complements information of the absent modality for single-modal clients.
%%
FedMEMA is validated on the BraTS 2020 benchmark for multimodal brain tumor segmentation.
Results show that it outperforms various up-to-date methods for multimodal and personalized FL and that its novel designs are effective.
Our code is available.
%The superiority of our proposed framework has been verified on (Brats2020) multimodal tumor brain image datasets, and ablation studies also demonstrate the effectiveness of the proposed method.
\end{abstract}

\section{Introduction}
\label{sec:intoduction}

Federated learning (FL) enables participants to collaboratively train a global model on their collective data without breaching privacy \cite{li2020federated}.
The decentralized mechanism makes it particularly suitable for privacy-sensitive application scenarios such as medical image analysis~\cite{kaissis2020secure,adnan2022federated,bercea2021feddis,experiments,yan2020variation}.
%{\color{green}where typical applications include
%More recently, Federated learning has made impressive achievements in a variety of medical tasks application, including 
%disease diagnosis~\cite{adnan2022federated,liu2020experiments,yan2020variation} and pathology  segmentation~\cite{bercea2021feddis}}.
However, most FL methods for medical image analysis only considered intramodal heterogeneity, limiting their applicability to multimodal imaging in practice.
%always focus on local uni-modal data, which limits their scalability in real-world application scenarios.
%%
%Different modalities of medical images can display information about different characteristics of human organs and diseased tissues. In clinical practice, in order to diagnose diseases more accurately and comprehensively, doctors often need to refer to multimodal image data for comprehensive analysis and judgment.

\begin{figure}[t]
\centering\footnotesize
\begin{tabular}{c m{0.8\columnwidth}}
     (a) & \includegraphics[width=.75\columnwidth]{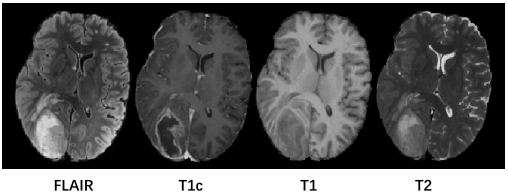}\\
     (b) & \includegraphics[width=.75\columnwidth]{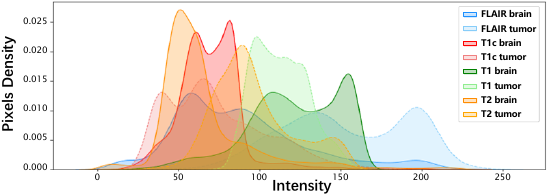}
\end{tabular}
\caption{(a) Example images of the four modalities in BraTS 2020 \cite{menze2014multimodal}.
(b) Histograms of the brain and tumor pixels for the four modalities.
}
\label{fig:3-mri}
\end{figure}

% \begin{figure}[htbp]
% \centering  %居中
% \subfigure[name of the subfigure]{   %第一张子图
% \begin{minipage}{7cm}
% \centering    %子图居中
% \includegraphics[scale=0.5]{sections/figs/mri.pdf}  %以pic.jpg的0.5倍大小输出
% \end{minipage}
% }
% \subfigure[name of the subfigure]{ %第二张子图
% \begin{minipage}{7cm}
% \centering    %子图居中
% \includegraphics[scale=0.8]{sections/figs/hist.pdf}%以pic.jpg的0.5倍大小输出
% \end{minipage}
% }
% \caption{Four modal MRI scans (a) from the BraTS2020 dataset together with histograms of the whole brain and the tumor pixels only (b).}    %大图名称
% \label{fig:3-mi}    %图片引用标记
% \end{figure}

One such application is brain tumor segmentation in multi-parametric magnetic resonance imaging (MRI) \cite{iv2018current}.
%{\color{green}---the current standard of care for clinical imaging diagnosis of brain tumors \cite{iv2018current}}.
Specifically, four MRI modalities (in this work, we refer to MRI sequences as modalities following literature \cite{dorent2019hetero,menze2014multimodal,shen2019brain,zhou2021latent}) are commonly used to provide complementary information and support sub-region analysis: T1-weighted (T1), contrast-enhanced T1-weighted (T1c), T2-weighted (T2), and T2 fluid attenuation inversion recovery (FLAIR), where the first two highlight tumor core and the last two highlight peritumoral edema (Fig. \ref{fig:3-mri}(a)).
%Fully aligned and paired multimodal MRI image data have been shown effective in brain tumor segmentation~\cite{pereira2016brain}.
When applying FL to such multimodal applications in practice, it is not uncommon that some participant institutes only possess a subset of the full modalities due to different protocols practiced,
%However, in real-world scenarios, some clinical institutions cannot collect full modal data due to equipment conditions and other factors.
%In medical scenarios, different medical institutions may not be able to obtain complete modal data due to the limitation of equipment conditions, and missing modality will inevitably lead to the degradation of local model performance. 
presenting a new challenge with the \textit{inter-modal heterogeneity} (Fig. \ref{fig:3-mri}(b)) across the participants of FL.
%This leads to data modalities that are not the same across institutions, introducing a new challenge - inter-modal heterogeneity.
%At this time, using federated learning to improve model performance will face a new challenge -- the heterogeneity caused by modal differences. 
In such a scenario, there can be two objectives for FL:
1) collectively training an optimal global model for full-modal input, and 
2) obtaining a personalized model for each participant \cite{chen2022fedmsplit,wang2019federated}, adapted for its data characteristics, and more importantly, better than trained locally without FL.
%the modalities they have locally by participating the FL.
To our knowledge, these two objectives were rarely considered together in FL for medical image analysis.

In this paper, we propose a new FL framework with federated modality-specific encoders and multimodal anchors (FedMEMA) for brain tumor segmentation.
%a multimodal FL framework that takes advantage of paired multimodal data on the server to solve the performance degradation problem of uni-modal clients, and allows clients to selectively learn useful knowledge from the global multimodal information through the attention mechanism to reduce the intra-modality heterogeneity caused by different modalities.
%%
Above all, to handle the distinctively heterogeneous MRI modalities, FedMEMA employs an exclusive encoder for each modality to allow a great extent of parameter specialization.
%we propose to use federated modality-specific encoders, with an exclusive encoder for 
%than the normalization parameters \cite{bernecker2022fednorm}.
In the meantime, while the encoders are shared between the server and clients, the decoders are personalized to cater to individual participants.
Specifically, a multimodal fusion decoder on the server (i.e., a participant with full-modal data) aggregates and fuses representations from the encoders to bridge the distribution gaps between modalities and reversely optimizes the encoders via backpropagation.
Meanwhile, multiple anchors are extracted from the fused multimodal representations and distributed to the clients in addition to the encoder parameters.
%who can access a small full-modal dataset for fusing multimodal features.
On the other end, the clients with incomplete modalities calibrate their local missing-modal representations toward the global full-modal anchors via the scaled dot-product attention mechanism \cite{vaswani2017attention} to make up the information loss due to absent modalities and adapt representations of present ones.
%further mitigating the impacts of inter-modal heterogeneity while adapting for local data.
%The calibration adaptively emphasizes parts of the global multimodal representations that best suits a client's data modality and distribution---via the dot attention mechanism \cite{vaswani2017attention}---to yield a model better tailored for the client.
To this end, we simultaneously obtain an optimal server model (for full-modal input) and personalized client models (for specific missing-modal input) from FL without sharing privacy-sensitive information.
%%
%The clients only need to share the feature extractor and utilize the cross-attention mechanism to complete local calibration with the clustering features returned by the server. 
%For the encoder parameters from clients, the corresponding modality is directly loaded, and then train and extract features on the global multimodal data.
%%

In summary, our contributions are as follows:
\begin{itemize}
    \item We bring forward the inter-modal heterogeneity problem due to missing modalities in FL for medical image analysis and aim to obtain an optimal full-modal server model and personalized missing-modal client models simultaneously with a novel framework coined FedMEMA.
    \item To tackle the inter-modal heterogeneity, we propose to employ a federated encoder exclusive for each modality followed by a server-end multimodal fusion decoder.
    Meanwhile, personalized decoders are employed for the clients to allow simultaneous personalization.
    \item In addition, we propose to extract and distribute multimodal representations from the server to the clients for local calibration of modality-specific features.
    \item Last but not least, we further enhance the calibration with multi-anchor representations.
\end{itemize}
Experimental results on the public BraTS 2020 benchmark show that our method achieves superior performance for both the server and client models to existing FL methods and that its novel designs are effective.
% \begin{itemize}
%     \item 
%      We propose a multimodal FL framework that works on clients with single modality and server with few multimodal data, which only transfers partial parameters and global feature anchors without revealing private data.
%     \item
%     In view of the multimodal data on the server, we propose a novel modal specific encoder and decoder architecture to replace the setting of a single encoder and decoder in traditional federated learning.
%     {\color{blue}Modality-specific encoders effectively utilize missing-modal data on clients to facilitate the FL (?).}
%     \item
%     The local calibration mechanism can calibrate the encoded features after encoder depending on the clustering features of the server, which allows each client to adaptively learn global multimodal information to better fit its local modal distribution.
%     % For the generation of global representation, the clustering method is used to slide update. The local calibration mechanism learns selectively through the attention mechanism.
%     \item
%     Our framework can enable single modal clients to absorb the modal-diversity knowledge through the knowledge transfer from multimodal server to local calibrate. At the same time, the sharing of feature extractor parameters also enables the server to obtain richer information of different modal implicitly.  
%     % In the final results, all participants achieved improved performance.
% \end{itemize}

\section{Related Work}

\subsubsection{Brain Tumor Segmentation with Multimodal MRI:}
%Multimodal brain tumor segmentation aims to use complete MRI data to achieve segmentation of different regions of brain tumors, which can help improve clinical diagnostic efficiency.
Multimodal MRI is the current standard of care for clinical imaging of brain tumors \cite{iv2018current}.
Segmentation and associated volume quantification of heterogeneous histological sub-regions are valuable to the diagnosis/prognosis, therapy planning, and follow-up of brain tumors \cite{menze2014multimodal}.
%Currently, most methods mainly focus on improving the architecture to achieve fusion of different modality features 
% reference; Brain tumor segmentation based on the fusion of deep semantics and edge information in multimodal MRI \cite{pereira2016brain,dolz2018hyperdense,zhang2021modality,liu2022sf,zhu2023brain}. 
In recent years, deep neural networks (DNNs) significantly advanced state-of-the-art of brain tumor segmentation with multimodal MRI \cite{chen2020brain,chen2019dual,ding2020multi,myronenko20183d,zhou2020one}.
%However, in practice, it is common to encounter missing modalities in brain tumor imaging due to image corruption, artifacts, acquisition protocols, contrast agent allergies, or simply cost constraints, making collecting complete modality data difficult and expensive.
However, these methods were optimized for ideal scenarios where the complete set of modalities was present. 
In practice, scenarios of missing one or more modalities commonly occur due to image corruption, artifacts, acquisition protocols, allergy to contrast agents, or cost.
%So recently there has been researches on brain tumor image segmentation with missing modalities \cite{hu2020knowledge,wang2021acn,ding2021rfnet,azad2022smu}.
Therefore, many efforts have been made to accommodate the practical scenarios of missing modalities \cite{hu2020knowledge,wang2021acn,ding2021rfnet,azad2022smu}.
These methods successfully improved DNNs' feature representation capability against missing modalities---however, only in the centralized setting, limiting their efficacy in privacy-sensitive settings.
%large amount of centralized data is required, otherwise performance will decline. However, this is difficult to achieve due to medical data privacy restrictions. Additionally, due to differences in equipment and resources between different institutions, it cannot be guaranteed that all institutions' data have complete modalities.
In this work, we aim to address the missing-modal problem in the FL setting and eliminate the privacy issue.

\subsubsection{Multimodal FL with Data Heterogeneity:}
%Heterogeneity is one of the most challenging issues in FL, as the models are trained locally on the client, different clients may have different computing capabilities, model architectures, and data distributions. Among these issues, data heterogeneity has attracted more attentions \cite{}. % citation and rewrite this sentence
%
Data heterogeneity is a primary challenge in FL \cite{mcmahan2017communication}.
%and has attracted more and more attentions in recent years. 
Personalized FL \cite{tan2022towards} proposed to adapt the global model locally on clients' data to address this issue.
%aims to address this challenge mainly by creating customized global models for each client \cite{tan2022towards}.
% reference: A state-of-the-art survey on solving non-IID data in Federated Learning  // Towards Personalized Federated Learning
However, it did not consider the heterogeneity due to multimodal data.
We are aware of several works for multimodal FL in the natural image domain.
%there has been limited research on the heterogeneity of data modality in FL with existing methods. 
%\cite{xiong2022unified} proposed a unified framework for multimodal federated learning, which used a co-attention mechanism to fuse complementary information from different modalities. 
%A personalization method based on MAML was used to adapt the final model for each client.
\citet{xiong2022unified} proposed a co-attention mechanism to fuse the complementary information of different modalities, yet requiring all clients to have access to the same set of modalities.
%\cite{zhao2022multimodal} proposed a multimodal FedAvg algorithm to aggregate local autoencoders trained on different modality combinations, higher weights will be assigned to multimodal clients, but this strategy requires manual weight adjustment and assumes the existence of multimodal clients in the FL framework. 
FedIoT \cite{zhao2022multimodal} employed cross-modal autoencoders to learn multimodal representations in an unsupervised manner.
However, both methods \cite{xiong2022unified,zhao2022multimodal} only obtained a single global classifier without catering to the personalized needs of modal-heterogeneous clients.
\citet{yu2023multimodal} proposed a cross-modal contrastive representation ensemble between the server and modal-heterogeneous clients by sharing a multimodal dataset, which may be unacceptable in strict privacy restrictions like healthcare.
In contrast, our framework optimizes a global model for full-modal input and simultaneously customizes a personalized model for each client's hetero-modal input.
It also maintains FL's data privacy by transmitting population-wise abstracted prototypes instead of image-wise features.

% donwei: the work below does not seem to be multimodal but multitasking to me.
%\citet{liu2020federated} extracted feature representations from different client \textbf{tasks} and aligning them on the server, which may violate the privacy protection of FL, as the directly encoded features can be used to reconstruct the original data, especially when using fixed pre-trained models.
%Another method~\cite{Liu2020FederatedLF} extracts representations from different modalities locally and sends them to the server to align different modalities. Representations encoded directly from the original data can be recovered, which raises privacy concerns, especially when the clients use a fixed pre-training model.

In medical image analysis, heterogeneity issues in multimodal FL have yet to be thoroughly discussed.
%\cite{chen2022fedmsplit} and \cite{yu2023multimodal} have focused on the natural scenario of modality incongruity, where clients have different sensors or are solving different tasks. 
%\cite{chen2022fedmsplit} proposed to divide the client model into smaller shareable blocks and allowing each type of block to provide a client-specific view of the relationship, which is passed through the graph. 
%Similarly, FedMsplit~\cite{chen2022fedmsplit} adaptively captures correlations between multimodal client models using a dynamic multi-view graph structure, but neighborhood message passing in the graph requires the expense of exposing model parameters to neighboring clients.
%However, this method requires the server to calculate the model correlation between each block of each client in each round, which is not feasible due to the high computational cost. \cite{yu2023multimodal} proposed a multimodal federated learning framework that enables training larger server models from clients with heterogeneous model architectures and data modalities, while only communicating features on public dataset. 
FedNorm \cite{bernecker2022fednorm} adapted the normalization parameters for different modalities while sharing common backbone parameters for CT- and MRI-based liver segmentation.
Yet, our experiments suggest that merely specializing in normalization parameters is insufficient to deal with the inter-modal heterogeneity in multimodal brain tumor segmentation.
%\cite{bernecker2022fednorm} proposed FedNorm and FedNorm+, two Federated Learning algorithms that use a modality-based normalization technique, successfully trained model on clients with data of high degree of heterogeneity (i.e. CT and MRI). 
%
In this work, we propose to handle the heterogeneity with modality-specific encoders to allow a greater extent of parameter specialization, followed by a multimodal fusion decoder to aggregate and fuse representations from the encoders and bridge the inter-modal distribution gaps.
%n this paper, we propose a new FL framework with federated modality-specific encoders and multimodal anchors (FedMEMA) for brain tumor segmentation. Although obtaining fully aligned multimodal data is challenging due to the differences in resources of different client devices, our assumption is that the server is a relatively resource-rich large medical institution, making it feasible to collect a certain amount of complete multimodal data. To handle different MRI modalities, we use modality-specific encoders for each modality, which have greater specialization than normalization parameters. Additionally, to compensate for information loss due to incomplete client modalities, while sharing encoder parameters, the server also passes on multimodal feature anchors for each category extracted from the multimodal fusion features, which are calibrated locally by the clients through a local calibration module to compensate for information loss. After fusion and compression of category features, the image feature representation has been sufficiently compressed and is difficult to be restored to data, which also alleviates privacy concerns in FL to some extent.

\begin{figure*}[t]
\centering
\includegraphics[width=.86\textwidth]{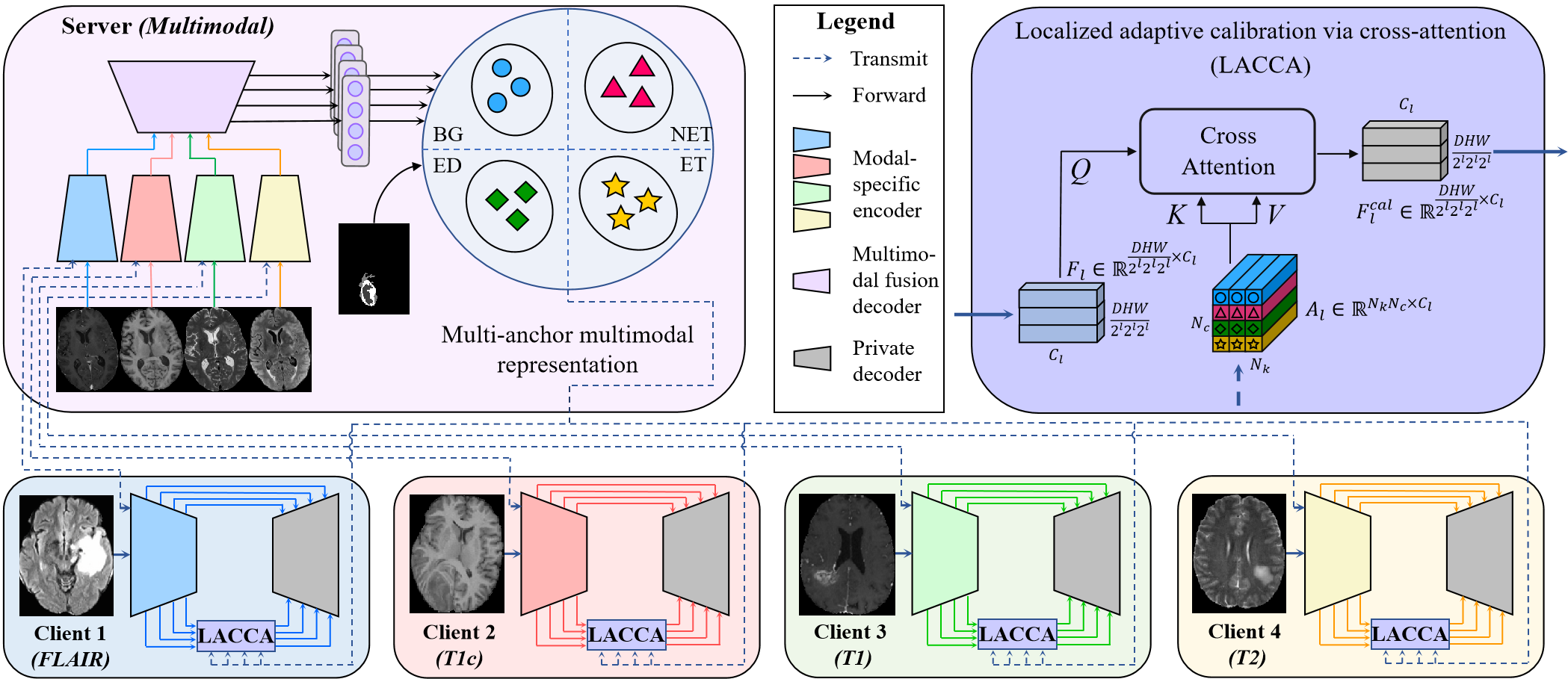} 
\caption{
 Overview of the proposed FedMEMA framework.
 FedMEMA employs a federated encoder exclusive for each modality followed by a server-end multimodal fusion decoder.
 Meanwhile, personalized decoders are used for the clients to allow simultaneous personalization.
 In addition, multi-anchor multimodal representations are extracted from the server and distributed to the clients for localized adaptive calibration of modality-specific features via cross-attention.
 ED: edema, ET: enhancing tumor, NET:  necrotic and non-enhancing tumor core, and BG: background.}
\label{fig:2-overview}
\end{figure*}

\section{Method}
% In this paper, we propose a multimodal FL framework 
%In this section, we present our proposed approaches for multimodal brain tumor segmentation on Magnetic Resonance Imaging (MRI) data with FL. Our goal is to enable FL to work on clients that have different local data modalities. Therefore, we first identify the application scenarios and some challenges. Finally, we introduce the details of our entire framework.

\subsubsection{Problem Definition:}
% {\color{green}In this work, we focus on the application of segmenting heterogeneous histological sub-regions of brain tumors in multi-parametric MRI, as an example of multimodal FL in medical image analysis.}
Let us denote the full set of modalities by $M=$\{T1, T1c, T2, FLAIR\} and a full-modal input by $X_M\in \mathbb{R}^{|M|\times D \times H\times W}$, where $D$, $H$, and $W$ are the depth, height, and width of the volume, respectively.
We consider a heterogeneous FL setting where a server with access to a set of full-modal data $\{(X_M, Y)\}$ coordinates several clients with access to data of incomplete modalities (``missing-modal''), where $Y \in \mathbb{R}^{N_c \times D \times H \times W}$ is the segmentation mask and $N_c$ is the number of target classes.
In practice, the server may be a major regional hospital, whereas the clients may be smaller local health units.
In this work, we mainly address the \textit{extreme and most challenging case} of monomodal clients, i.e., each client houses data of a specific modality denoted by $\{(X_m, Y)\}$, where $X_m\in\mathbb{R}^{1\times D\times H\times W}$, for a pioneer and exploratory methodology development
(some preliminary results in the more common settings of multimodal clients are provided in the supplementary material).
Furthermore, we mainly consider the horizontal FL setting where 1) each client has different patients, or 2) a client may share some patients with another, i.e., various clients may have different modalities of the same patients.
The latter scenario is practical as a patient may have different imaging studies at different hospitals within a district.
%{\color{red}On the contrary, it is not practical to assume all clients have the same patients.}
However, considering the privacy issue, the clients do not know whether or which of their data overlaps with others'.

This work aims to train, via FL, not only an optimal global model that works well with full-modal data but also optimal personalized models that work well in specific missing-modal situations for the clients.
The latter is practically meaningful as it is usually difficult for a local health unit to upgrade its imaging protocol quickly, yet still look forward to training an optimal model for its current protocol by participating in the FL.

% We consider a heterogeneous FL setting where $N$ uni-modal image clients cooperate with server that has some multimodal data to achieve overall performance improvement, while obtaining a global modal that adapts to different modal inputs.
% %%
% Each image client has its own private dataset $N_n={(x_n^i,y_n^i)}_{i=1}^{|N_n|}$, where $x_n^i$ is the $i$-th training sample of the $n$-th client, $y_n^i$ is the corresponding annotation.

\subsubsection{Framework Overview:}
Without loss of generality, we use four clients, each with data of a mutually different modality, for method description.
As shown in Fig.~\ref{fig:2-overview}, the server has four modality-specific encoders (one for each modality) and a modal fusion decoder, whose fused features are clustered to produce multimodal anchors.
Meanwhile, each client has a modality-specific encoder for local data modality and a private decoder.
Additionally, a localized adaptive calibration via cross-attention (LACCA) module calibrates the clients' missing-modal representations toward the server's multimodal anchors.
%{\color{green}Below, we describe each component and the FL process in detail.}
%{\color{red}In each training round, the clients receive encoder parameters and the multimodal anchors from the server, where the latter is used to calibrate local features extracted by the former via a LACCA (localized adaptive calibration via cross-attention) module during client-end training.
%After local training, the clients transmit the encoder parameters to the server, where the parameters and multimodal anchors are updated via server-end training, and distributed back for the next round.}
%during training, the clients first receive global categoy representations from the multimodal paired data as well as the encoder parameters of the corresponding modality, and then perform the local calibration in the decoding stage with the cross-attention module (Section~\ref{2-2}). Finally the clients transmit the local encoder parameters to the server.
%To avoid the possible privacy leakage problem of directly transmitting the representations, the server uses clustering to generate multiple multimodal representations for each category (Section~\ref{2-3}), and then sends them and the trained modal encoder parameters back to the clients.

% \subsection{Global Training and Category Multimodal Representation Generation}
\subsubsection{FL with Modality-specific Encoders:}
In classical FedAvg \cite{mcmahan2017communication}, the server and clients usually share the same network architecture, where the server aggregates and averages the network parameters of the clients and then distributes the averaged parameters back to the clients in a straightforward manner.
However, due to the high heterogeneity among the multimodal MRI data (see Fig.~\ref{fig:3-mri}),
%of the clients in our problem setting: data of different modalities at different clients,
%, i.e. clients with only one MRI imaging modality, 
%the task of multimodal brain tumor segmentation with FL is 
our problem setting becomes challenging for this paradigm.
%\footnote{Implementation-wise, it is still possible to apply the FedAvg paradigm, where most of the network architectures are the same for the server and clients with only the number of input channels adapted.
%However, as we will show in the experiments, this solution yields apparently inferior performance to our proposed method.}
%contains multiple challenges.
%%
 %%
%2) The data distribution of different clients is unbalanced.
%% (写上的话实验应该需要考虑clients之间数量的不均衡?)
%3) The scanns are performed with different scanners and imaging protocols, resulting in different image resolutions, sizes and inter-slice spacing. 
%%
%In addition, 
%For example, the four MRI different imaging modalities have different visual appearances and pixel features, see in Fig.~\ref{fig:3-mri}. %% 像素强度直方图
%Rather than adopting a universe network architecture for the server and clients, 
Instead, we propose federated modality-specific encoders to handle the distinctively heterogeneous imaging modalities.
On the one hand, we adopt an 
%encoder-decoder 
architecture with late fusion strategy \cite{ding2021rfnet} to compose the global model on the server, including a modality-specific encoder $E_m$ (parameterized by $W_m^s$) for each modality, a fusion decoder $D_M$ for multimodal feature aggregation and fusion,  and a regularizer (not shown in Fig. \ref{fig:2-overview} for simplicity). 
The regularizer is a straightforward auxiliary segmentation decoder shared by all modality-specific encoders.
It regularizes the encoders to learn the same discriminative
features by forcing them to share the decoder parameters.
Please refer to \cite{ding2021rfnet} for details.\footnote{Note that our framework is model-agnostic and can be implemented with various non-FL multimodal segmentation models consisting of modality-specific encoders and modal fusion decoder(s) \cite[e.g.,][]{dorent2019hetero,shen2019brain,zhou2021latent}.
In this work, we use \cite{ding2021rfnet} for demonstration due to its outstanding performance and straightforward architecture.}
%but to demonstrate the efficacy of our framework for medical multimodal FL scenarios.
Given a full-modal input $X_M$, each $E_m$ first extracts features from the corresponding modality $X_m$, followed by $D_M$ fusing multimodal features and generating segmentation masks.
On the other hand, each single-modal client has a federated modality-specific encoder $E_m$ and a personalized decoder $D_m$.
% for generating segmentation masks
$E_m$ on the clients shares the same architecture as the server.

In each round of FL, the clients first receive parameters $W_m^s$ from the server to replace its local copy $W_m^i$, where $i\in\{1,\ldots, N_m\}$ and {\color{red}$N_m$ is the number of clients with data of modality $m$}, train for $N_e$ epochs on local data, and then send updated $W_m^i$ back to the server.
After receiving $W_m^i$, the server averages $W_m^i$ of the same modality (if $N_m>1$): $W_m^s = \frac{1}{N_m}\sum_i W_m^i$, train for $N_e$ epochs on the full-modal data, and sends updated $W_m^s$ to the clients for the next round.\footnote{We are aware of strategies for dynamic weight assignment for parameter aggregation based on clients’ data sizes \citep[e.g.,][]{hsu2020federated}.
Although this work focuses on inter-modal heterogeneity and assigns the same amount of data to all clients, we expect the incorporation of dynamic weight assignment to make our method more robust in practice.}
Thus, the server bridges the distribution gaps between modalities with the fusion decoder $D_M$ and utilizes the complementary multimodal information to train each modality-specific encoder $E_m$ via backpropagation.

\subsubsection{Multi-Anchor Multimodal Representation:} %  Generation
\label{2-2}
Besides aligning modality-specific encoders with the multimodal fusion decoder, the server also generates multi-anchor multimodal representations for the classes of interest, which will be distributed to the clients in addition to the encoder parameters.
%for client-end calibration of missing-modal features toward the full-modal distributions and representation power,
%(to describe in the next section), 
%thus mitigating the data heterogeneity issue.
\citet{liu2020federated} proposed communicating encoded representations, which may breach the privacy restriction.
% %To tackle the problem of federated heterogeneity, some articles adopt the approach of communicating on encoded features and prototypes. li2021model
On the contrary, some works proposed to transmit category prototypes \cite{mu2023fedproc,tan2022fedproto}.
%instead of raw features for transmission efficiency
Yet a single prototype was highly compressed and may not carry enough representative information for a class, especially considering the significant inter-subject variations in 3D multimodal medical images.
%\textbf{Are existing prototypes multimodal? And there is no multi-anchor ones?}
%%
%The method of transmitting prototypes has been studied in image classification tasks~\cite{tan2022fedproto,mu2023fedproc}, but the prototypes compress the information, so that the valuable information is too little, which cannot achieve good performance in image segmentation task, especially in 3D.
%% 3D切patch也会造成特征空间维度的不一致性

In this work, we propose to extract \textit{multiple} prototypes \cite{cui2020unified} from the fused multimodal features for each class of interest for enhanced representation power, which we refer to as anchors for their calibration purpose \cite{ning2021multi}.
% potential multimodal distribution
%%Therefore, we learn a multimodal mixture representation for each category (comprising several prototypes instead of a single prototype).
%%
{\color{red}Concretely, we extract per-class features from the fused multimodal feature maps of the decoder $D_M$ by masked average pooling using the ground truth mask and apply the K-means method \cite{macqueen1967classification} to the extracted features to obtain $N_k$ anchors.}
Intuitively, these $N_k$ anchors are the modes of each class's multimodal distribution.
%By concatenating the features of each category into an image-level vector, we perform clustering on them to estimate the distribution of each category.
For the $l$\textsuperscript{th} feature scale level, where $l\in\{1,\ldots,4\}$ for the networks we use, the anchors for all the $N_c$ classes can be collectively denoted by $A_l\in\mathbb{R}^{N_k N_c \times C_l}$, where $C_l$ is the number of feature channels.
Empirically, we determine the cluster membership using the most abstract feature level, i.e., $l=4$, and apply the membership to compute $N_k=3$ anchors for all levels (see corresponding experiments in the next section).
This strategy can well preserve the full-modal information of each class while incurring little network transmission burden. % than raw features
It should also be noted that the few class-wise anchors are abstracted from the entire training population on the server, thus carrying little privacy information concerning individuals.
To avoid the collapse of the training process due to jumps in cluster centroids as a result of re-clustering at each round \cite{xie2016unsupervised}, we treat the anchors as a memory bank and update them smoothly via exponential moving average~\cite[EMA;][]{tarvainen2017mean}: $\Bar{a}_c = \omega\Bar{a}_c + (1-\omega)a_c$, 
%\begin{equation}
%    \Bar{a}_c = \omega\Bar{a}_c + (1-\omega)a_c,
%\end{equation}
where $\Bar{a}_c$ is an anchor for class $c$ in the memory bank and updated by the closest cluster centroid $a_c$ of the same class, and $\omega$ is set to 0.999 following \citet{tarvainen2017mean}.

%{\color{blue}These full-modal anchors are distributed to the missing modal clients, where they are used to calibrate the local features towards full-modal representation capability.}

\begin{algorithm}[!t]
    \caption{FedMEMA algorithm. Note that the personalized (non-federated) decoder parameters are not shown below for simplicity.}
    % $\beta$ is the smoothing coefficient of EMA, 
    \begin{algorithmic}[1]\small
    \Require the modality set $M$ = \{T1, T1c, T2, FLAIR\} indexed by $m$, a full-modal training dataset $\mathcal{D}_M$ on the server, the number of clients $N_m$ with data of modality $m$, the monomodal training set $\mathcal{D}_m^i$ on client $i$ with data of modality $m$,
    %$A_l$ is the global multi-anchor multimodal representation for $l^{th}$ feature scale level where $l\in\{1,\ldots,4\}$,  
    the number of communication rounds $N_r$, and the number of training epochs $N_e$ in each round.
    %and the smoothing coefficient $\alpha$ for exponential moving average (EMA).
    \renewcommand{\algorithmicensure}{\textbf{Output}:} 
    \Ensure the collection of parameters $W^s_{\{m\}}=\{W^s_m\}$ for the modality-specific encoders,
    and the collection of multimodal anchors $A_{\{l\}}=\{A_l\}$ for different feature scale levels $l$ (plus parameters of the personalized decoders).
    \vspace{1ex}
        \renewcommand{\algorithmicfunction}{\textbf{Server executes}:} 
        \Function{}{}
        \State Initialize $W_{\{m\}}^s$, and update $W_{\{m\}}^s$ by training on $\mathcal{D}_M$ for $N_e$ epochs
        \State Initialize $A_{\{l\}}$ by K-means
        \For {round $r=1$ to $N_r$}
            \For {$m \in M$}
            \For {each client $i\in N_m$}
            \State $W_m^i \gets$ ClientUpdate$(m, i, W_m^{s}, A_{\{l\}})$ \Comment{run on client $i$}
            % \STATE $W_m \gets W_m \cup W_{t+1}^{m_k}$
            \EndFor
            \State $W_m^s = \frac{1}{N_m}\sum_i W_m^i$ \Comment{aggregate parameters for modality-specific encoder}
            \EndFor
            \State Update $W_{\{m\}}^s$ by training on $\mathcal{D}_M$ for $N_e$ epochs
            \State Update $A_{\{l\}}$ by exponential moving average
            %EMA with the hyper-parameter $\alpha$
            % \STATE $A_{t+1}^l = \beta A_{t}^l + (1-\beta )\cdot {\theta}_{t+1}$
            % \State $A_{t+1}^l$ = EMA$(A_{t+1}^l)$
        \EndFor\EndFunction\vspace{1ex}

        \renewcommand{\algorithmicfunction}{\textbf{ClientUpdate}($m,i,W^s_m,A_{\{l\}}$):} 
        \Function{}{}\Comment{run on client $i$ with modality $m$}
        \State $W_m^i \gets W_m^s$
        % \For{epoch $e=1, \ldots, N_e$}
        % \State $W_m^{i} \gets W_m^{i} - \eta \nabla \mathcal{L}(b, A_l; W^{i})$
        \State Update $W_m^{i}$ by training on $\mathcal{D}_m^i$ with LACCA (Eq. (\ref{eq})) for $N_e$ epochs
        % \EndFor
        \State return $W_m^i$
        \EndFunction
    \end{algorithmic}
\end{algorithm}

\begin{table*}[t]
\centering
\begin{adjustbox}{width=.81\width}
\begin{tabular}{c|cccccc}
\hline
Number of anchors ($N_k$) & FLAIR          & T1c            & T1              & T2             & Avg            & S              \\ \hline
1                 & 60.45\scriptsize{$\pm$16.85}$^*$ & 75.34\scriptsize{$\pm$22.02}$^*$ & 54.73\scriptsize{$\pm$19.29} & 57.12\scriptsize{$\pm$14.90}$^*$ & 61.91\scriptsize{$\pm$14.74}$^*$ & 83.80\scriptsize{$\pm$16.26}$^*$ \\
2                 & 60.22\scriptsize{$\pm$15.61}$^*$ & 77.29\scriptsize{$\pm$21.86}$^*$ & 57.66\scriptsize{$\pm$18.21}$^*$ & 59.08\scriptsize{$\pm$15.71}$^*$ & 63.56\scriptsize{$\pm$14.63}$^*$ & 83.91\scriptsize{$\pm$17.23}$^*$ \\
\rowcolor[HTML]{EFEFEF}3       & \textbf{62.52}\scriptsize{$\pm$16.79} & 76.37\scriptsize{$\pm$21.83} & 57.26\scriptsize{$\pm$17.24} & \textbf{60.36}\scriptsize{$\pm$17.23} & \textbf{64.13}\scriptsize{$\pm$15.17} & \textbf{84.17}\scriptsize{$\pm$17.19} \\
4                 & 61.16\scriptsize{$\pm$16.87}$^*$ & 76.71\scriptsize{$\pm$22.44}$^*$ & 56.28\scriptsize{$\pm$16.56}$^*$ & 59.67\scriptsize{$\pm$16.47}$^*$ & 63.45\scriptsize{$\pm$14.93}$^*$ & 83.85\scriptsize{$\pm$15.90}$^*$ \\
5                 & 59.91\scriptsize{$\pm$17.72}$^*$ & \textbf{77.37}\scriptsize{$\pm$25.25}$^*$ & 57.21\scriptsize{$\pm$16.97}$^*$ & 59.05\scriptsize{$\pm$16.70}$^*$ & 63.38\scriptsize{$\pm$16.09}$^*$ & 83.95\scriptsize{$\pm$15.73}$^*$ \\
7                 & 60.93\scriptsize{$\pm$17.83}$^*$ & 76.69\scriptsize{$\pm$22.55}$^*$ & \textbf{57.84}\scriptsize{$\pm$18.21}$^*$ & 58.93\scriptsize{$\pm$18.63}$^*$ & 63.60\scriptsize{$\pm$15.64}$^*$& 83.46\scriptsize{$\pm$16.17}$^*$ \\ \hline
\end{tabular}
\end{adjustbox}

% \vspace{1mm}

% \setlength{\tabcolsep}{.8mm}
\begin{adjustbox}{width=.81\width}
\begin{tabular}{c|cccccc}
\hline
{\phantom{ii}Feature scale level ($l$)\phantom{iii}} & FLAIR          & T1c             & T1              & T2             & Avg            & S              \\ \hline
{1}                         & 59.58\scriptsize{$\pm$16.33}$^*$ & 77.06\scriptsize{$\pm$20.52}$^*$ & 56.63\scriptsize{$\pm$18.46}$^*$ & 58.68\scriptsize{$\pm$14.52}$^*$ & 62.99\scriptsize{$\pm$14.55}$^*$ & 84.05\scriptsize{$\pm$16.88}$^*$ \\
{2}                         & 61.15\scriptsize{$\pm$16.54}$^*$ & 76.69\scriptsize{$\pm$23.49}$^*$ & 56.12\scriptsize{$\pm$18.37}$^*$ & 58.38\scriptsize{$\pm$16.67}$^*$ & 63.08\scriptsize{$\pm$15.74}$^*$ & 83.93\scriptsize{$\pm$16.73} \\
{3}                         & 60.26\scriptsize{$\pm$15.78}$^*$ & \textbf{77.10}\scriptsize{$\pm$22.03}$^*$ & 56.99\scriptsize{$\pm$18.61}$^*$ & 58.77\scriptsize{$\pm$15.70}$^*$ & 63.28\scriptsize{$\pm$15.43}$^*$ & 83.21\scriptsize{$\pm$16.19}$^*$ \\
\rowcolor[HTML]{EFEFEF}4                  & \textbf{62.52}\scriptsize{$\pm$16.79} & 76.37\scriptsize{$\pm$21.83} & \textbf{57.26}\scriptsize{$\pm$17.24} & \textbf{60.36}\scriptsize{$\pm$17.23} & \textbf{64.13}\scriptsize{$\pm$15.17} & \textbf{84.17}\scriptsize{$\pm$17.19} \\
{1--4}                      & 60.11\scriptsize{$\pm$15.37}$^*$ & 75.99\scriptsize{$\pm$23.95}$^*$ & 56.67\scriptsize{$\pm$18.12} & 59.45\scriptsize{$\pm$16.03}$^*$ & 63.05\scriptsize{$\pm$15.42}$^*$ & 83.41\scriptsize{$\pm$16.47}$^*$ \\ \hline
\end{tabular}
\end{adjustbox}
\caption{Results of experimental setting 1 on the \textit{validation} set in mDSC (\%).
Top: varying $N_k$ (number of multimodal anchors per class) with $l=4$. 
Bottom: varying the feature scale level $l$ (with $N_k=3$) of the multimodal fusion decoder $D_M$, based on which the cluster membership is determined;
$l=4$ indicates the most abstract level of the smallest scale (i.e., at the bottleneck between the encoders and decoder), and ``1--4'' concatenates features of all four levels together for clustering.
FLAIR, T1c, T1, and T2 indicate performance of the clients with the corresponding data modalities, Avg indicates their average, and ``S'' indicates server performance.
%and overall indicates the average performance across all sites (including both the server and clients).
*: $p<0.05$ comparing against $N_k=3$ (top) and $l=4$ (bottom), respectively, in each column.}%
\label{tabs:clusNum}
\end{table*}

\subsubsection{Localized Adaptive Calibration via Cross-Attention:}\label{2-3}
%the parameters $W_m^s$ corresponding to their respective modalities plus
In each federated round, the clients receive from the server the multimodal anchors $A_l$,  
%where the former is used to replace the local parameters and 
which are used to calibrate local missing-modal representations.
%%At each federated round, all clients receive multimodal hybrid representations of each category and the encoder parameters $E_m$ of their corresponding modality from global server in the last round, while the parameters of prediction head and attention module are {\color{red}initialized} from local training themselves. 
%%
Concretely, denoting the final feature map at the $l$\textsuperscript{th} scale level of the encoder by $F_l\in\mathbb{R}^{\frac{D}{2^l} \times \frac{H}{2^l} \times \frac{W}{2^l} \times C_l}$,
%, where $C_l$ is the channel number, 
we reshape $F_l$ to the dimension $\frac{DHW}{2^l2^l2^l} \times C_l$.
Then, inspired by the attention operation in the Transformer architecture \cite{vaswani2017attention}, we treat the reshaped $F_l$ as queries and the multimodal anchors as the keys and values, and calibrate the local representations toward the global multimodal anchors by the cross attention:
% {\color{red}(add the projection: are there projections and how?)}
\begin{equation}\label{eq}
    F_l^\mathrm{cal} = \operatorname{Attn}(F_l, A_l)=\operatorname{softmax}\big({F_l A_l^T}\big/{\sqrt{C_l}}\big)A_l.
\end{equation}
%%
%% 本地模型更新公式,参考personal FL
% \begin{equation}
% \end{equation}
%Concretely, in the $n$-th local client, the encoded feature of a input data at the $l$-th encoding stage is denoted as $F_n^l\in \mathbb{R}^{C\times \frac{D}{2^l} \times \frac{H}{2^l} \times \frac{W}{2^l}}$, where $C$ is the channel number and $(D,H,W)$ are the image size and $l=4$ in this work.
%%
% As the core of the Transformer architecture, the cross-attention is formulated as:
% \begin{equation}
%     Attn(Q, K, V) = softmax(QK^\top / \sqrt{d})V,
% \end{equation}
% where $Q, K, V$ are sets of query, key, and value vectors, and $d$ is the dimension of the query and key vectors.
%%
%%In this work, we adopy Eq.(1) to compute adaptation coefficient.
%%
%Locally encoded features $F_n^l\in \mathbb{R}^{C\times \frac{D}{2^l} * \frac{H}{2^l} {\ast }  \frac{W}{2^l}}$ as query vectors, and global category multimodal hybrid representations $glbF\in \mathbb{R}^{k\times cls*C}$ as key and value vectors. 
Finally, the calibrated features $F_l^\mathrm{cal}$ are reshaped back and element-wise added to the final features of the same scales of the decoder to participate in the subsequent forward propagation.
The calibration process is localized and self-adaptive in that each client locally emphasizes the part of the global multi-anchor multimodal representations that best suits its own data modality and distributions---via the dot-product attention---to yield more powerful models tailored for itself.
%Since attention modules LCA is private locally, they can adaptively calibrate feature representation learning to better fit their own data modality distribution.
%For the global multimodal hybrid representations of each category, we propose a Cross-Attention-based Local Calibration mechanism (LCA) to adaptively emphasize the information better suited for the current modality, rather than treating all modes equally or relying on only one of them for local calibration.
To this end, we name it the \textit{localized adaptive calibration via cross-attention} (LACCA) module.
The LACCA module is inserted in all four feature scales of our backbone networks.
Note that the multimodal anchors are learnable parameters during FL and are directly used by the clients for inference after training.
%Each client updates the model for optimization using its local uni-modal data ${(x_n^i,y_n^i)}_{i=1}^{|N_n|}$ and global hybrid multimodal representations $glbF$.
The complete algorithm of our method is detailed in Algorithm 1.

\begin{table*}[t]
\centering

\begin{adjustbox}{width=.755\textwidth}
\begin{tabular}{c|cccccc}
\hline
\multirow{2}{*}{Method} & \multicolumn{6}{c}{Setting 1}                                                                                                                                                                                                                 \\ \cline{2-7} 
                        & FLAIR                                 & T1c                                   & T1                                    & T2                                    & Avg                                   & S                                     \\ \hline
Local models            & 55.21\scriptsize{$\pm$18.57}$^*$      & 64.60\scriptsize{$\pm$25.13}$^*$      & 42.98\scriptsize{$\pm$18.83}$^*$      & 54.84\scriptsize{$\pm$16.78}$^*$      & 54.41\scriptsize{$\pm$16.22}$^*$      & 79.91\scriptsize{$\pm$16.48}$^*$      \\
RFNet                   & 54.65\scriptsize{$\pm$18.07}$^*$      & 67.70\scriptsize{$\pm$22.89}$^*$      & 43.69\scriptsize{$\pm$16.92}$^*$      & 56.49\scriptsize{$\pm$16.44}$^*$      & 55.63\scriptsize{$\pm$15.43}$^*$      & 79.46\scriptsize{$\pm$17.63}$^*$      \\ \hline
FedAvg                  & 54.04\scriptsize{$\pm$17.48}$^*$      & 62.40\scriptsize{$\pm$24.57}$^*$      & 36.60\scriptsize{$\pm$14.71}$^*$      & 54.11\scriptsize{$\pm$14.56}$^*$      & 51.78\scriptsize{$\pm$14.17}$^*$      & 78.42\scriptsize{$\pm$17.23}$^*$      \\
PerFL                   & 54.94\scriptsize{$\pm$15.42}$^*$      & 63.56\scriptsize{$\pm$26.13}$^*$      & 39.08\scriptsize{$\pm$15.25}$^*$      & 53.57\scriptsize{$\pm$15.09}$^*$      & 52.78\scriptsize{$\pm$14.53}$^*$      & 78.71\scriptsize{$\pm$16.80}$^*$      \\
FedNorm                 & 57.06\scriptsize{$\pm$17.04}$^*$      & 68.45\scriptsize{$\pm$21.13}$^*$      & 39.90\scriptsize{$\pm$17.60}$^*$      & 56.16\scriptsize{$\pm$16.56}          & 55.39\scriptsize{$\pm$15.64}$^*$      & 77.72\scriptsize{$\pm$17.44}$^*$      \\
CreamFL                 & 55.07\scriptsize{$\pm$16.42}$^*$      & 66.24\scriptsize{$\pm$25.66}$^*$      & 47.01\scriptsize{$\pm$18.30}$^*$      & 56.92\scriptsize{$\pm$18.48}$^*$      & 56.31\scriptsize{$\pm$16.01}$^*$      & 80.30\scriptsize{$\pm$16.48}$^*$      \\
FedMSplit               & 56.54\scriptsize{$\pm$17.45}$^*$      & 69.58\scriptsize{$\pm$22.56}$^*$      & 46.31\scriptsize{$\pm$16.99}$^*$      & 59.96\scriptsize{$\pm$15.20}$^*$      & 58.09\scriptsize{$\pm$14.75}$^*$      & 81.01\scriptsize{$\pm$15.93}$^*$      \\
FedIoT                  & 58.85\scriptsize{$\pm$17.51}$^*$      & 71.14\scriptsize{$\pm$21.01}$^*$      & 49.96\scriptsize{$\pm$17.62}$^*$      & 61.07\scriptsize{$\pm$15.29}          & 60.25\scriptsize{$\pm$15.22}$^*$      & 80.23\scriptsize{$\pm$16.76}$^*$      \\ \hline
FedMEMA (ours)          & \textbf{61.02}\scriptsize{$\pm$16.12} & \textbf{71.91}\scriptsize{$\pm$23.32} & \textbf{53.56}\scriptsize{$\pm$18.06} & \textbf{61.55}\scriptsize{$\pm$15.54} & \textbf{62.01}\scriptsize{$\pm$15.12} & \textbf{82.11}\scriptsize{$\pm$18.04} \\ \hline
\end{tabular}
\end{adjustbox}

\begin{adjustbox}{width=.755\textwidth}
\begin{tabular}{c|cccccc}
\hline
\multirow{2}{*}{Method} & \multicolumn{6}{c}{Setting 2}                                                                                                                                                                                                                 \\ \cline{2-7} 
                        & FLAIR                                 & T1c                                   & T1                                    & T2                                    & Avg                                   & S                                     \\ \hline
Local models            & 55.35\scriptsize{$\pm$19.06}$^*$      & 69.67\scriptsize{$\pm$22.11}$^*$      & 44.24\scriptsize{$\pm$19.87}$^*$      & 57.47\scriptsize{$\pm$15.56}$^*$      & 56.68\scriptsize{$\pm$15.73}$^*$      & 79.91\scriptsize{$\pm$16.48}$^*$      \\
RFNet                   & 54.65\scriptsize{$\pm$18.07}$^*$      & 67.70\scriptsize{$\pm$22.89}$^*$      & 43.69\scriptsize{$\pm$16.92}$^*$      & 56.49\scriptsize{$\pm$16.44}$^*$      & 55.63\scriptsize{$\pm$15.43}$^*$      & 79.46\scriptsize{$\pm$17.63}$^*$      \\ \hline
FedAvg                  & 55.59\scriptsize{$\pm$17.14}$^*$      & 65.47\scriptsize{$\pm$23.11}$^*$      & 42.65\scriptsize{$\pm$18.84}$^*$      & 56.61\scriptsize{$\pm$18.22}$^*$      & 55.08\scriptsize{$\pm$17.00}$^*$      & 78.19\scriptsize{$\pm$16.83}$^*$      \\
PerFL                   & 56.70\scriptsize{$\pm$16.09}$^*$      & 64.24\scriptsize{$\pm$23.22}$^*$      & 45.16\scriptsize{$\pm$17.01}$^*$      & 57.43\scriptsize{$\pm$14.71}$^*$      & 55.88\scriptsize{$\pm$15.73}$^*$      & 80.09\scriptsize{$\pm$16.74}          \\
FedNorm                 & 55.78\scriptsize{$\pm$17.24}$^*$      & 70.91\scriptsize{$\pm$22.11}$^*$      & 50.75\scriptsize{$\pm$17.76}$^*$      & 51.80\scriptsize{$\pm$16.12}$^*$      & 57.31\scriptsize{$\pm$15.47}$^*$      & 78.28\scriptsize{$\pm$17.44}$^*$      \\
CreamFL                 & 59.98\scriptsize{$\pm$16.56}$^*$      & 69.54\scriptsize{$\pm$23.70}$^*$      & 50.05\scriptsize{$\pm$17.97}$^*$      & 59.55\scriptsize{$\pm$17.16}$^*$      & 59.78\scriptsize{$\pm$16.26}$^*$      & 81.55\scriptsize{$\pm$15.22}$^*$      \\
FedMSplit               & 58.87\scriptsize{$\pm$16.26}$^*$      & 70.70\scriptsize{$\pm$23.08}$^*$      & 50.41\scriptsize{$\pm$17.10}$^*$      & 60.11\scriptsize{$\pm$16.45}$^*$      & 60.02\scriptsize{$\pm$15.09}$^*$      & 80.86\scriptsize{$\pm$16.33}          \\
FedIoT                  & 60.47\scriptsize{$\pm$15.07}$^*$      & 71.96\scriptsize{$\pm$21.20}$^*$      & 52.49\scriptsize{$\pm$17.28}          & 61.03\scriptsize{$\pm$14.83}$^*$      & 61.48\scriptsize{$\pm$14.64}          & 81.47 \scriptsize{$\pm$17.31}$^*$     \\ \hline
FedMEMA (ours)          & \textbf{62.84}\scriptsize{$\pm$16.17} & \textbf{73.49}\scriptsize{$\pm$21.77} & \textbf{56.46}\scriptsize{$\pm$19.04} & \textbf{61.58}\scriptsize{$\pm$14.87} & \textbf{63.59}\scriptsize{$\pm$15.27} & \textbf{83.27}\scriptsize{$\pm$17.27} \\ \hline
\end{tabular}%\vspace{-.5mm}
\end{adjustbox}
\caption{Experimental results on the \textit{test} set in mDSC (\%).
FLAIR, T1c, T1, and T2 indicate the clients' performance with the corresponding data modalities, Avg indicates their average, and ``S'' indicates server performance.
*: $p<0.05$ comparing against our method in each column.}
\label{tabs:state-of-art}
\end{table*}

\section{Experiments and Results}

\subsubsection{Dataset and Experimental Settings:}
We conduct experiments on the multimodal Brain
Tumor Segmentation (BraTS) 2020 dataset~\cite{menze2014multimodal,bakas2018identifying}, which consists of 369 multi-contrast MRI scans with four sequences: T1, T1c, T2, and FLAIR.
The goal is to segment three nested subregions of brain tumors: whole tumor, tumor core, and enhancing tumor.
Following \citet{ding2021rfnet}, we divide the dataset into 219, 50, and 100 subjects for training, validation, and testing, respectively,
%% 取验证集最好的epoch
The test set is used only for the final model evaluation, whereas the validation set is used for model optimization.
Without loss of generality, we design two experimental settings per our problem definition.\footnote{\color{blue}In this paper, we primarily focus on the inter-modal heterogeneity due to missing modalities but ignore the intramodal heterogeneity due to institutions, i.e., potential distribution discrepancies between data of the same modality but from different institutions.}
In setting 1, the training set is evenly divided among the server and four clients at random, i.e., no subject overlap between clients.
Thus, the server and clients include 43, 44, 44, 44, and 44 training subjects, respectively.
In setting 2, the server data remain unchanged, whereas the rest of the training data are randomly divided into $4+1$ equal parts where the additional ``1'' is the common data for all clients.
The server and clients include 43, 71, 71, 71, and 71 training subjects, respectively.
% {\color{red}During division, we ensure all cases collected from the same institute according to the official record of BraTS 2020 are assigned to the same ``part''.}
Meanwhile, the server and all clients in both settings use the same validation and test sets.
%In addition, we vary the number of modalities available to each client from one to three in each setting, {\color{blue}and set the number of clients such that every possible combination of the available modalities is entertained by one client.}
Note that each client can only access the specific modalities of its assigned data.
%the modality of four clients is completely different and half the data is paired). 
%Setting 2 can simulate a situation where a patient may visit more than one hospital and a record exists.
The mean of the Dice similarity coefficients (mDSC) of the three tumor subregions is employed as the evaluation metric, and the Wilcoxon signed rank test is used to analyze statistical significance.

%%
%~\cite{}
% Dice coefficient is used to measure the segmentation performance of the proposed method, defined as:
% \begin{equation}
%     Dice_{cls}(y, \hat{y}) = \frac{2\cdot {\left \| y_{cls}\cap \hat{y}_{cls} \right \|}_1 }{{\left \| y_{cls} \right \|}_1 + {\left \| \hat{y}_{cls} \right \|}_1 }
% \end{equation}
% Where $cls$ denotes different tumor classes. $Dice_{cls}$ denotes the Dice Score of the tumor class $cls$, $y$ and $\hat y$ correspond to ground truth and annotation respectively. 
%After selecting the model that performs best on the validation set, we evaluate the final performance on the test set corresponding to each site. 
%{\color{red}Do we need to mention our simulation setting? How about related works?}

\subsubsection{Implementation:}
The proposed FL framework is implemented using PyTorch (1.13.0) and trained with five RTX 2080Ti GPUs, with the server on one GPU and the clients evenly distributed on the rest.
We use the RFNet \cite{ding2021rfnet} as our server network, and its modality-specific encoder and regularizing decoder as the clients' encoder and decoder, respectively.
The LACCA module is implemented with eight attention heads.
%{\color{red}The detailed architectures of the networks we use are provided in the supplementary material.}
%The smoothing coefficient of EMA is set to 0.999 following \citet{tarvainen2017mean}.
The input crop size is $80\times80\times80$ voxels, and the batch size is set to 1 and 3 for the server and clients, respectively.
Other settings are the same for the server and clients.
The commonly used Dice loss \cite{milletari2016v} plus the cross entropy loss for medical image segmentation are employed.
The Adam optimizer, with its learning rate and weight decay set to 0.0002 and $10^{-5}$, respectively, is leveraged for optimization.
We train the networks for 1000 rounds, and in each federated round, the server and clients are trained for one epoch.
We follow \citet{ding2021rfnet} for data preprocessing and augmentation.
Our code is available at https://github.com/QDaiing/FedMEMA.
%{\color{green}For reproducible research, our implementation, including code and data split, will be published with the paper.}

\begin{figure*}[t]
\centering
\includegraphics[width=.89\textwidth]{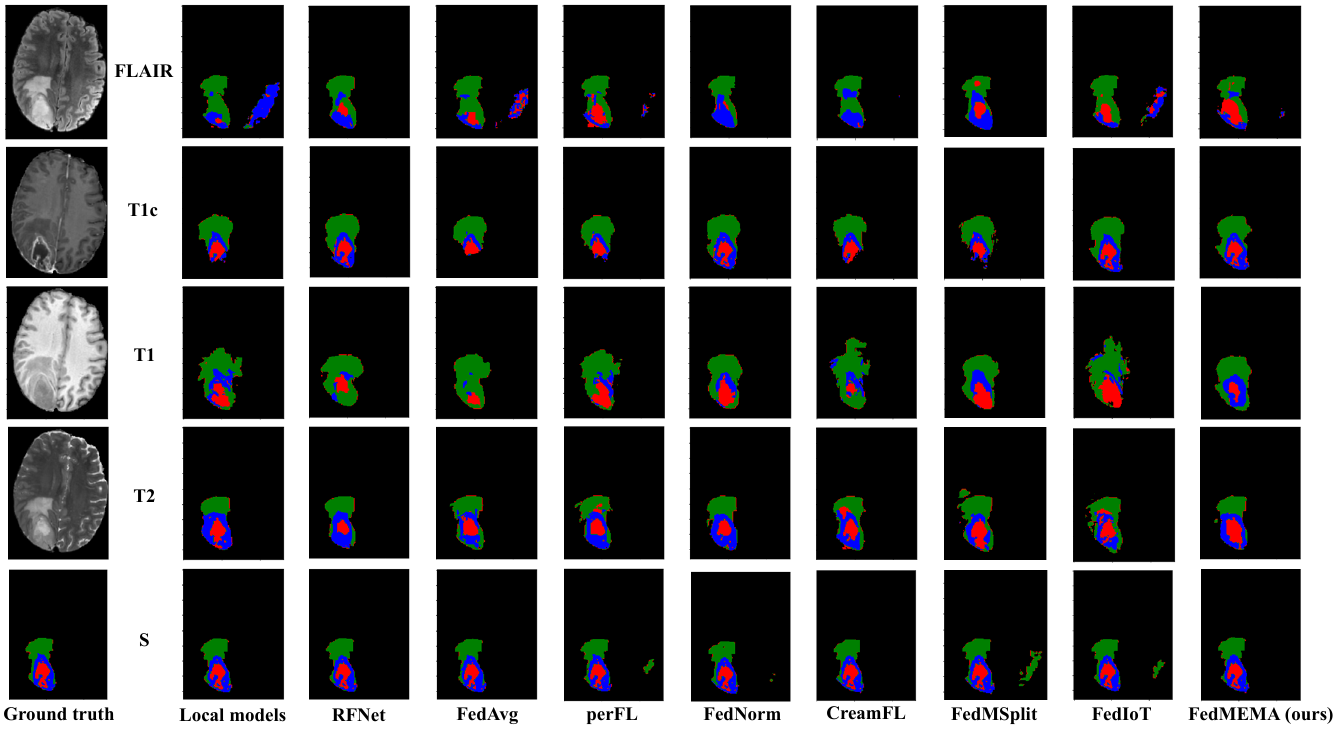} 
\caption{Example segmentation results in experimental setting 1 for a subject in the test set.
%, compared to those by several state-of-the-art methods.
FLAIR, T1c, T1, and T2 indicate the clients with the corresponding data modalities, and ``S'' indicates the server.
Red: necrotic and non-enhancing tumor core, blue: enhancing tumor, and green: edema.
% Comparisons with the state-of-the-art in Setting 1 on Brats 2020 dataset.
}
\label{fig:Setting1}
\end{figure*}

\begin{table*}[t]
\centering
\setlength{\tabcolsep}{1mm}
\begin{adjustbox}{width=.87\textwidth}
\begin{tabular}{lccc|cccccc}
\hline
Ablation & Server & Federated & LACCA        & FLAIR          & T1c            & T1             & T2             & Avg            & S              \\ \hline
% (a)      & E\&D   & E\&D      &              & 54.54          & 69.06          & 40.71          & 51.92          & 54.06          & 79.75          \\
(a)      & E\&D   & D         & -            & 43.81\scriptsize{$\pm$17.12}$^*$ & 48.83\scriptsize{$\pm$25.13}$^*$ & 18.34\scriptsize{$\pm$9.72}$^*$ & 43.59\scriptsize{$\pm$18.10}$^*$ & 38.64\scriptsize{$\pm$11.25}$^*$ & 80.05\scriptsize{$\pm$15.46}$^*$ \\
(b)      & E\&D   & E         & -            & 51.48\scriptsize{$\pm$15.62}$^*$ & 58.51\scriptsize{$\pm$17.17}$^*$ & 36.35\scriptsize{$\pm$12.47}$^*$ & 47.97\scriptsize{$\pm$17.06}$^*$ & 48.58\scriptsize{$\pm$11.27}$^*$ & 81.11\scriptsize{$\pm$16.38}$^*$ \\ \hdashline
(c)  & 4E\&D  & 4E        & -            & 57.14\scriptsize{$\pm$16.42}$^*$ & 72.95\scriptsize{$\pm$14.66}$^*$ & 50.67\scriptsize{$\pm$16.13}$^*$ & 58.01\scriptsize{$\pm$17.12}$^*$ & 59.69\scriptsize{$\pm$12.84}$^*$ & 82.11\scriptsize{$\pm$16.42}$^*$ \\
(d)  & 4E\&D  & 4E        & Mono-anchor  & 60.45\scriptsize{$\pm$18.07}$^*$ & 75.34\scriptsize{$\pm$22.89}$^*$ & 54.73\scriptsize{$\pm$16.92} & 57.12\scriptsize{$\pm$16.44}$^*$ & 61.91\scriptsize{$\pm$15.43}$^*$ & 83.80\scriptsize{$\pm$17.63}$^*$ \\ \hline
(e) Ours     & 4E\&D  & 4E        & Multi-anchor & \textbf{62.52}\scriptsize{$\pm$16.00} & \textbf{76.37}\scriptsize{$\pm$14.56} & \textbf{57.26}\scriptsize{$\pm$15.09} & \textbf{60.36}\scriptsize{$\pm$18.39} & \textbf{64.13}\scriptsize{$\pm$12.84} & \textbf{84.17}\scriptsize{$\pm$11.54} \\ \hline
\end{tabular}%\vspace{-.5mm}
\end{adjustbox}
% \begin{tabular}{cccc|cccccc}
% \toprule
% \multicolumn{4}{c|}{\multirow{2}{*}{lc glb shareP LCA}} & \multicolumn{6}{c}{Average DCS (\%) $\uparrow$}             \\
% \multicolumn{4}{c|}{}                                   & A     & B     & C     & D     & Avg     & S     \\ \hline
% E\&D & E\&D & \color{red}E\&D \\
% E\&D         & E\&D         & D        & -              & 42.03 & 43.62 & 17.16 & 47.29 & 37.52  & 78.13 \\
% E\&D         & E\&D         & E        & -              & 51.02 & 58.51 & 32.63 & 50.03 & 48.05 & 78.16 \\
% E\&D         & 4E\&D        & 4E        & -              & 55.44 & 69.17 & 49.34 & 57.64 & 57.89 & 80.74 \\
% E\&D         & 4E\&D        & 4E        & proto        & 58.81 & 71.14 & 48.84 & 60.5  & 59.82 & 81.41 \\
% E\&D         & 4E\&D        & 4E        & clusF         & \textbf{61.02} & \textbf{71.91} & \textbf{53.56} & \textbf{61.55} & \textbf{62.01}   & \textbf{82.11} \\ 
% \bottomrule
% \end{tabular}
\caption{Ablation study on the \textit{validation} set in experimental setting 1 using mDSC (\%).
``E'' and ``D'' are encoder and decoder, respectively.
FLAIR, T1c, T1, and T2 indicate the clients' performance with the corresponding data modalities, Avg indicates their average, and ``S'' indicates server performance.
*: $p<0.05$ comparing against our method in each column.}%
%"lc" and "glb" denote the model architecture of local and global, and "shareP" means the parameters shared in FL. "proto" means prototype and "clusF" means clustering features.
\label{tabs:ablation study}
\end{table*}

\subsubsection{Validating Designs for Multi-Anchor Multimodal Representations:}
Based on the validation data, we first determine 1) the optimal number of multimodal anchors per class ($N_k$) and 2) the optimal feature scale level $l$ based on which the cluster membership is determined.
We fix either of them to reduce the search space while varying the other.
The results are shown in Table \ref{tabs:clusNum}.
Although the results look fairly stable, we select $N_k=3$ and $l=4$ for evaluation and comparison with other methods on the test data due to their highest performances in both the clients' average and the server's mDSCs compared with alternative values.
We conjecture that $l=4$ (i.e., the most abstract feature level) works the best due to its great capability of abstraction and denoising despite the relatively low resolution.

\subsubsection{Comparison with Baseline and State-of-the-Art (SOTA) Methods:}
We compare our proposed FedMEMA to various baseline and SOTA FL algorithms.
As the baseline, the server and client models are trained locally on the private data of each site.
%, whose performance can be regarded as the lower bound
%Additionally, we train one centralized model on the collection of training data from all participants, which can be seen as upper bounds.
As to SOTA FL algorithms, we adopt the classical FedAvg \cite{mcmahan2017communication} and several up-to-date approaches to FL on multimodal data or personalized models, including FedNorm \cite{bernecker2022fednorm}, FedMSplit \cite{chen2022fedmsplit}, CreamFL \cite{yu2023multimodal}, FedIoT \cite{zhao2022multimodal}, and PerFL~\cite{wang2019federated}.
%clusterFL~\cite{qayyum2022collaborative}
Under the premise of keeping the methodological principles unchanged, necessary adaptations are made to ensure fair comparison:
%in our practical experimental settings: 
%As the practical problem definition we consider in medical scenarios in this work were rarely considered in previous work, 
%different approaches, which tackle the problem of multimodal data heterogeneity in FL and have a similar setting,
%%
1) for FedAvg and its derived methods (i.e., PerFL and FedNorm), which did not originally conduct server-end training, we make them do so on the server data as our method,\footnote{Our preliminary experiments empirically showed that with the server-end training, they performed better than without it.} and 
2) we change FedIoT's autoencoding clients to supervised networks.
Also, we use the same networks as our clients' for FedAvg and derivatives (in FedAvg infrastructure, the server and clients use the same networks) and the same networks as our server and clients for the counterparts in other methods.
It should be noted that as CreamFL requires sharing the server data with all clients, %for contrastive representation ensemble, 
it violates the privacy restriction in the medical context and increases the training data for the clients.
%{\color{green}Therefore, its results should be referenced with due caution.}
Lastly, as RFNet \cite{ding2021rfnet} was originally designed for both full- and missing-modal segmentation after training with full-modal data, we also train it on the server data in its original recipe and evaluate its performance for comparison.
% \textbf{More comparisons!?}}
%For implementation, since these methods are originally designed for different tasks, we try our best to keep their design principle and adapt them to our federated setting and 3D brain tumor image segmentation task.

The results are shown in Table~\ref{tabs:state-of-art}.
As the comparative trends are similar in both settings, we mainly describe setting 1 below.
%%
%As we can see, the performance of the RFNet trained on the server data is similar to the baseline performance of training separate models 
Without modality-specific parameters, FedAvg and PerFL mostly yield worse performance than the baseline local models.
This indicates that the inter-modal heterogeneity impedes the classical FL from effectively utilizing the extra data on the clients.
%It can be seen that in the setting where the client modalities are not identical to each other, the introduction of the federated learning paradigm does not bring performance improvement due to higher heterogeneity.
By specializing the normalization parameters for different modalities, FedNorm achieves slight improvements in average mDSC across the clients but with slight to modest decreases in the server's performance.
%%
%In comparison, sharing extracted feature and personalizing the projection layers (PerFL) are slightly useful for local performance improvement, which is also the core idea of the design of our own framework. 
%%
The more complex and advanced CreamFL, FedMSplit, and FedIoT frameworks can achieve comprehensive improvements over the baseline;
especially, FedIoT exceeds in the clients' average mDSC by close to 6\%.
% Simple personalized federated learning algorithm that fine-tune the aggregated model on local dataset only provide a slight improvement (1.0\%) and still a gap with Local Train.
%FedIoT and CreamFL are adaptive in our setting but not significant, with 3-4\% improvement over Local Train. 
%This may be because our all clients are single modal, while their performance improvement depends more on the participation of multimodal clients, which is demonstated in their original experiments.
%%
%For clusterFL and FedNorm whithout the help of auxiliary multimodal data on the server, the improvement is smaller, especially less than 1\% in FedNorm. After careful experiments, we find that the Mode Normalization (MN) technique proposed by FedNorm is not very suitable on our dataset, causing a certain performance loss, which may be the reason for its poor performance.
In comparison, our FedMEMA further improves upon FedIoT by $\sim$2\% in both the clients' average and server mDSCs and achieves the best performance for all clients and the server.
We attribute our method's advantages to the adaptive calibration via cross-attention between local monomodal and global full-modal representations.
Remarkably, besides improving the clients' performance with locally personalized models, FedMEMA also substantially enhances the server's performance by effectively exploiting the clients' data of heterogeneous modalities.
%%
%By introducing global multimodal data and assisting the client to perform local calibration, our method achieves optimal performance on all metrics, especially 7.6\% increase on averaged Dice score and 2.2\% increase on the server.
Meanwhile, the RFNet, despite being a robust model for various missing-modal situations, yields performance comparable to the baseline local models, probably because it does not use the extra data on clients.
Lastly, the performance in setting 2 is generally better than that in setting 1, likely due to the more training data available on each client.
Fig. \ref{fig:Setting1} shows example segmentation results.
%%% 脑肿瘤分割图的可视化及分析

\subsubsection{Ablation Study:}
%To prove the effectiveness of each critical component, we perform ablation studies and the results are presented in Table~\ref{tabs:ablation study}.
We conduct thorough ablation studies to validate the efficacy of our novel framework design, including the federated modality-specific encoders and personalized decoders, the LACCA module, and the multi-anchor multimodal representations.
The results are presented in Table~\ref{tabs:ablation study}, where the first two rows are variants of the classical FedAvg \cite{mcmahan2017communication} with either the encoder or decoder federated, and the last three are variants of our proposed method.
It can be seen that, among the variants of FedAvg, federating the encoder while personalizing the decoder (row (b)) outperforms the reverse (row (a)).
%The first two rows are our baselines, local clients and server follow the same Encoder-Decoder architecture, the first row shares the feature extraction layer - Encoder, and the second row shares the prediction layer - Decoder. In contrast, the shared representation achieves better performance.
%%
On top of that, our modality-specific encoders (row (c)) achieve substantial performance boosts over the FedAvg family, e.g., the clients' average mDSC improves by $\sim$11\%
%and the server's mDSC improves by {\color{red}x.xx}\% 
compared with row (b), demonstrating the effectiveness of the FL architecture in our problem setting.
Row (d) additionally incorporates the LACCA module but with mono-anchor representations obtained by averaging all server data, achieving further improvements, especially on the clients ($>$2\% in average mDSC).
This suggests that multimodal representations are effective for the adaptive calibration of local monomodal features.
%However, here the features are compressed to the prototypes, maybe loss some information, so the improvement is not as significant as our method. 
%%
The last row is our full model with multi-anchor representations, yielding the best performances in both average missing-modal mDSC across the clients ($\sim$64\%) and full-modal mDSC on the server ($\sim$84\%).
These results indicate that multiple anchors represent the 3D multimodal medical data better than the mono ones, as expected.
%achieving the average Dice gains of around 14\%, and global model improvement of 3.95\%. 
%%
%Thanks to the multimodal paired data and mode-specific encoders architecture on the server, as well as the local calibration mechanism based on global class multimodal features, each prediction layer on the client can adaptively emphasize the feature space of its own modality to better mark each tumor.

\section{Conclusion}
In this paper, we proposed FedMEMA---a new FL framework with federated modality-specific encoders and multimodal anchors for brain tumor segmentation in multi-parametric MRI with missing-modal clients, 
%FedMEMA tackled the inter-modal heterogeneity by 1) employing a federated encoder exclusive for each modality followed by a server-end multimodal fusion decoder, 2) extracting and distributing multimodal representations to the clients for local calibration of modality-specific features, and 3) further enhancing the calibration with multi-anchor representations.
%%
%The proposed framework can improve the performance of all uni-modal clients by transmitting global multimodal features and presentation layer parameters without leaking patient privacy and data. And because of the superiority of the global model, we additionally obtain a network that can adapt to different modal inputs, and its performance is also competitive.
%%
and demonstrated its superior performance to existing FL methods on the public BraTS 2020 benchmark in the extreme case of monomodal clients.
%Experimental results on the public BraTS 2020 benchmark demonstrated FedMEMA's superior performance to existing FL methods.
%, validating the efficacy of modality-specific encoders and multimodal anchors in addressing modality heterogeneity.
%In this paper, we mainly focus on the modality heterogeneity issue and the other types of heterogeneity are out of our scope, which is the limitation of this paper.
%In the future, we will study the heterogeneity of data distribution and model architecture in federated learning.
In future work, we plan to evaluate FedMEMA with the more common settings of hetero-modal clients and more datasets.
In addition, this paper mainly focused on the inter-modal heterogeneity due to missing modalities but ignored the intramodal heterogeneity due to institutions, which was a limitation.
We also plan to consider both types of heterogeneities together in the future.

\section{Acknowledgments}
This work was supported in part by the National Key R\&D Program of China under Grant 2020AAA0109500/ 2020AAA0109501, and in part by the National Natural Science Foundation of China (Grant No. 62371409).

\newpage

\setcounter{page}{1}
\setcounter{figure}{0}
\setcounter{table}{0}
\makeatletter
\renewcommand{\thepage}{S\arabic{page}}
\renewcommand{\thefigure}{S\arabic{figure}}
\renewcommand{\thetable}{S\arabic{table}}
\renewcommand{\thealgorithm}{S\arabic{algorithm}}
\begin{strip}

\centering
\begin{adjustbox}{width=.9\textwidth}
\begin{tabular}{c}
     \textbf{\textit{Supplementary Material}: Federated Modality-specific Encoders and}\\
     \textbf{Encoders and Multimodal Anchors for Personalized Brain Tumor Segmentation}\\
\end{tabular}
\end{adjustbox}

\vspace{20mm}
\begin{center}\vspace{-12mm}
\begin{minipage}{\textwidth}
\setlength{\tabcolsep}{.5mm}
\centering
\begin{adjustbox}{width=0.8\textwidth}
\begin{tabular}{ccccccccccccccccccccc}
\hline
\multirow{2}{*}{Methods}             & \multicolumn{6}{c}{Whole tumor}                      &  & \multicolumn{6}{c}{Tumor core}                       &  & \multicolumn{6}{c}{Enhancing tumor}                  \\ \cline{2-7}\cline{9-14}\cline{16-21}
  & FLAIR   & T1c   & T1    & T2    & Avg   & S && FLAIR   & T1c   & T1    & T2    & Avg   & S & & FLAIR   & T1c   & T1    & T2    & Avg   & S \\ \hline
Local models    & 79.84 & 61.39 & 64.14      & 78.12 & 70.87 &   87.15                 
&  & 54.07          & 68.87  &  43.32    & 54.89   
&55.29& \underline{82.46} &    &31.73& 63.55 & 21.48 & 31.51 & 37.07 & 70.13    \\
RFNet    & 79.40 & 66.78 & 65.06      & 80.25 & 72.87 &   87.01                 
&  & 56.42          & 72.89  &  45.14    & 57.69   
&58.04& 81.92 &    &28.11& 63.44 & 20.87 & 31.53 & 35.99 & 69.44     \\ \hline
FedAvg    & 79.09 & 64.13 & 60.73      & 78.14 & 70.52 &   86.54                 
&  & 52.52          & 64.88  &  38.96    & 55.24   
&52.90& 78.59 &    &30.52& 58.19 & 10.12 & 28.96 & 31.94 & 70.13     \\
PerFL    & 81.03 & 63.30 & 55.94      & 76.68 & 69.24 &   86.86                 
&  & 53.44          & 66.74  &  42.02    & 53.72   
&53.98& 79.90 &    &30.34& 60.65 & 19.27 & 30.31 & 35.14 & 69.38     \\
FedNorm   & 80.50 & 65.29 & 52.63      & 76.99 & 68.85 &   85.64                 
&  & 57.49          & 70.86  &  44.81    & 57.71   
&57.72& 79.43 &    &31.73& 63.55 & 21.48 & 31.51 & 37.07 & 70.13     \\
CreamFL    & 82.07 & 64.91 & \underline{68.81}      & 80.59 & 74.10 &   87.99                 
&  & 54.25          & 69.15  &  48.17    & 55.68   
&56.81& 82.05 &    &33.19& \underline{69.21} & 22.25 & 33.78 & 39.61 & 70.85    \\
FedMSplit    & \underline{82.15} & 68.14 & 68.33      & \underline{82.37}& 75.25 &   \underline{88.69}                 
&  & 56.67          & 73.25  &  48.80    & 58.68   
&56.81& 82.05 &    &30.81& 67.34 & 21.82 & \underline{35.85} & 38.96 & \underline{72.03}    \\
FedIoT    & 82.03 & \underline{70.98} & 68.47      & 81.28 & \underline{75.69} &   87.71                 
&  & \underline{59.86}          & \textbf{76.50}  &  \underline{52.68}    & \textbf{63.59}   
&\underline{63.16}& 81.50 &    &\underline{34.67}& 65.93 & \underline{28.72}& \textbf{38.34} & \underline{41.92} & 71.49    \\ \hline
Ours    & \textbf{84.26} & \textbf{71.49} & \textbf{71.37}      & \textbf{82.88} & \textbf{77.50} &   \textbf{89.48}                 
&  & \textbf{61.39}          & \underline{74.37}  &  \textbf{57.23}    & \underline{63.43}   
&\textbf{64.11}& \textbf{84.29} &    &\textbf{37.41}& \textbf{69.87} & \textbf{32.09} & \textbf{38.34} & \textbf{44.43} & \textbf{72.57}     \\
%%%%%%%%%%%%%%%%%%%
%\rowcolor[HTML]{EFEFEF}
%\multicolumn{4}{c}{Win / 15}                    &  & 0    & 1    & 2             & \textbf{7}     & 6 &                & 0    & 2             & 0     & 3        & \textbf{10}    &     & 0             & 1 & 1 & 2 & \textbf{12}                     \\
%\multicolumn{4}{c}{$p$ value}                    &  & $<$$10^{-4}$ & $<$$10^{-4}$ & $<$0.001 & -                     &  & $<$$10^{-4}$ & $<$$10^{-4}$ & 0.19 & -                     &  & $<$$10^{-4}$ & $<$$10^{-4}$ & 1.00             & -                     \\ 
\hline
\end{tabular}
\end{adjustbox}
\captionof{table}{Class-wise results in mDSC (\%) of different tumor subregions for the setting-1 results shown in Table 2.
The local models are trained locally on the private data of each site.
FLAIR, T1c, T1, and T2 indicate the clients' performance with the corresponding data modalities, Avg indicates their average, and ``S'' indicates server performance.}
%
%*: $p<0.05$ comparing against our method in each column (NOT mandatory).}% by Wilcoxon signed rank test
\label{tab:subregion}
\end{minipage}\\
\vspace{5mm}
\begin{minipage}{\textwidth}
\setlength{\tabcolsep}{.5mm}
\centering
\begin{adjustbox}{width=0.9\textwidth}
\begin{tabular}{c|cccccccccc}
\hline
% \multirow{2}{*}{Method} & \multicolumn{10}{c}{8 clients}   \\  \cline{2-11} 
Method & FLAIR\textsuperscript{1}   & T1c\textsuperscript{1}   & T1\textsuperscript{1}   & T2\textsuperscript{1}   & FLAIR\textsuperscript{2}   & T1c\textsuperscript{2}   & T1\textsuperscript{2}    & T2\textsuperscript{2}    & Avg   & S          \\ \hline
Local models    & 40.66\scriptsize{$\pm$16.60}$^*$ & 38.47\scriptsize{$\pm$16.76}$^*$ & 41.97\scriptsize{$\pm$16.33}$^*$ & 42.14\scriptsize{$\pm$18.83}$^*$ & 36.12\scriptsize{$\pm$15.99}$^*$ & 38.52\scriptsize{$\pm$18.47}$^*$ & 38.39\scriptsize{$\pm$17.90}$^*$ & 35.88\scriptsize{$\pm$15.31}$^*$ & 39.02\scriptsize{$\pm$15.29}$^*$ & 79.91\scriptsize{$\pm$16.48}$^*$ \\
% RFNet \\ \hline
RFNet  & 54.65\scriptsize{$\pm$18.07}$^*$ & \textbf{67.70}\scriptsize{$\pm$22.89}$^*$  & 43.69\scriptsize{$\pm$16.92}$^*$  & 56.49\scriptsize{$\pm$16.44}$^*$ & \underline{54.65}\scriptsize{$\pm$18.07}$^*$ & \underline{67.70}\scriptsize{$\pm$22.89}$^*$ & 43.69\scriptsize{$\pm$16.92}$^*$ & \underline{56.49}\scriptsize{$\pm$16.44}$^*$ & \underline{55.63}\scriptsize{$\pm$15.43}$^*$ & 79.46\scriptsize{$\pm$17.63}$^*$ \\ \hline
FedAvg  & 51.29\scriptsize{$\pm$18.28}$^*$ & 56.04\scriptsize{$\pm$28.51}$^*$ & 32.78\scriptsize{$\pm$15.72}$^*$ & 51.45\scriptsize{$\pm$15.99}$^*$ & 49.90\scriptsize{$\pm$17.95}$^*$ & 59.48\scriptsize{$\pm$25.72}$^*$ & 34.04\scriptsize{$\pm$15.49}$^*$ & 50.04\scriptsize{$\pm$15.99}$^*$ & 48.13\scriptsize{$\pm$14.40}$^*$ & 76.31\scriptsize{$\pm$17.88}$^*$ \\
PerFL  & 50.24\scriptsize{$\pm$18.37}$^*$ & 55.54\scriptsize{$\pm$29.23}$^*$ & 33.49\scriptsize{$\pm$14.87}$^*$ & 51.70\scriptsize{$\pm$15.70}$^*$ & 52.41\scriptsize{$\pm$16.80}    & 59.54\scriptsize{$\pm$26.74}$^*$ & 34.62\scriptsize{$\pm$14.79}$^*$ & 51.88\scriptsize{$\pm$16.82}$^*$ & 48.68\scriptsize{$\pm$15.12}$^*$ & 75.66\scriptsize{$\pm$16.35}$^*$ \\
FedNorm & 49.81\scriptsize{$\pm$14.63}$^*$ & 59.84\scriptsize{$\pm$27.32}$^*$ & 31.19\scriptsize{$\pm$14.06}$^*$ & 46.95\scriptsize{$\pm$14.29}$^*$ & 49.87\scriptsize{$\pm$15.07}$^*$ & 59.99\scriptsize{$\pm$28.55}$^*$ & 28.63\scriptsize{$\pm$13.28}$^*$ & 49.55\scriptsize{$\pm$15.34}$^*$ & 46.98\scriptsize{$\pm$13.11}$^*$ & 76.38\scriptsize{$\pm$18.83}$^*$ \\
CreamFL & \underline{56.23}\scriptsize{$\pm$16.96} & 65.46\scriptsize{$\pm$24.33}$^*$ & 43.27\scriptsize{$\pm$15.99}$^*$ & 53.39\scriptsize{$\pm$16.59}$^*$ & 50.57\scriptsize{$\pm$16.66}$^*$ & 63.45\scriptsize{$\pm$26.66}$^*$ & 40.59\scriptsize{$\pm$17.02}$^*$ & 50.83\scriptsize{$\pm$16.83}$^*$ & 52.97\scriptsize{$\pm$14.34}$^*$ & 80.40\scriptsize{$\pm$15.62}$^*$ \\ 
FedMSplit & 45.30\scriptsize{$\pm$16.22}$^*$    & 63.64\scriptsize{$\pm$27.78}$^*$ & \underline{45.29}\scriptsize{$\pm$16.18}$^*$  & 50.73\scriptsize{$\pm$14.48}$^*$ & 44.08\scriptsize{$\pm$15.50}$^*$ & 63.86\scriptsize{$\pm$26.73}$^*$ & 43.96\scriptsize{$\pm$16.54}     & 55.41\scriptsize{$\pm$14.07}$^*$ & 51.53\scriptsize{$\pm$14.18}$^*$ & 80.96\scriptsize{$\pm$16.51}$^*$ \\
FedIoT   & {55.18}\scriptsize{$\pm$16.69}$^*$    & 67.55\scriptsize{$\pm$23.51}    & 44.56\scriptsize{$\pm$15.95}$^*$ & \underline{58.17}\scriptsize{$\pm$14.42}$^*$ & 52.52\scriptsize{$\pm$16.22}    & 67.39\scriptsize{$\pm$25.92}$^*$    & \textbf{47.13}\scriptsize{$\pm$18.12}$^*$     & 50.60\scriptsize{$\pm$16.81}$^*$    & 55.39\scriptsize{$\pm$15.10}$^*$    & \underline{81.77}\scriptsize{$\pm$16.32}$^*$    \\ \hline
Ours   & \textbf{56.81}\scriptsize{$\pm$15.72}      & \underline{67.65}\scriptsize{$\pm$25.96}      & \textbf{46.48}\scriptsize{$\pm$14.84}       & \textbf{59.04}\scriptsize{$\pm$14.82}       & \textbf{54.78}\scriptsize{$\pm$17.36}       & \textbf{68.45}\scriptsize{$\pm$24.38}        & \underline{47.05}\scriptsize{$\pm$15.46}       & \textbf{58.29}\scriptsize{$\pm$16.53}       & \textbf{57.32}\scriptsize{$\pm$14.55}       & \textbf{82.31}\scriptsize{$\pm$16.74}          \\ \hline
\end{tabular}
\end{adjustbox}
\captionof{table}{Experimental results on the \textit{test} set in mDSC (\%) with \textit{eight clients}.
The local models are trained locally on the private data of each site.
The compared methods are RFNet \cite{ding2021rfnet}, FedAvg \cite{mcmahan2017communication}, PerFL~\cite{wang2019federated}, FedNorm \cite{bernecker2022fednorm}, CreamFL \cite{yu2023multimodal}, FedMSplit \cite{chen2022fedmsplit}, and FedIoT \cite{zhao2022multimodal}.
FLAIR\textsuperscript{\{1,2\}}, T1c\textsuperscript{\{1,2\}}, T1\textsuperscript{\{1,2\}}, and T2\textsuperscript{\{1,2\}} indicate the performance of the clients with the corresponding data modalities, Avg indicates their average, and ``S'' indicates server performance.
%c1-8 indicate performance of the clients while c1 and c5, c2 and c6, c3 and c5, c4 and c8 correspond to FLAIR, T1c, T1, T2 respectively.
*: $p<0.05$ comparing against our method in each column.}% by Wilcoxon signed rank test
\label{tab:8_clients}
\end{minipage}
\end{center}
\end{strip}

\begin{table*}[!t]
\centering
\begin{adjustbox}{width=1.4\columnwidth}
\begin{tabular}{ccccccc}
\hline
Bimodal & Local models                     & FedAvg                           & CreamFL                          & FedMSplit                        & FedIoT                           & Ours                             \\ \hline
Avg & 64.12\scriptsize{$\pm$15.42}$^*$ & 63.38\scriptsize{$\pm$14.83}$^*$ & 64.02\scriptsize{$\pm$15.53}$^*$ & 69.81\scriptsize{$\pm$15.34}$^*$ & 65.12\scriptsize{$\pm$14.93}$^*$ & \textbf{70.61}\scriptsize{$\pm$15.23} \\
S   & 79.91\scriptsize{$\pm$16.48}$^*$ & 78.92\scriptsize{$\pm$23.38}$^*$ & 80.53\scriptsize{$\pm$22.30}$^*$ & 79.75\scriptsize{$\pm$22.11}$^*$ & {80.75}\scriptsize{$\pm$22.83} & \textbf{81.07}\scriptsize{$\pm$23.53} \\ \hline
\hline
Trimodal & Local models                     & FedAvg                           & CreamFL                          & FedMSplit                        & FedIoT                           & Ours                             \\ \hline
Avg & 72.45\scriptsize{$\pm$16.16}$^*$ & 71.74\scriptsize{$\pm$15.16}$^*$ & 73.23\scriptsize{$\pm$17.15}$^*$ & {77.5}\scriptsize{$\pm$15.30} & 75.49\scriptsize{$\pm$15.36}$^*$& \textbf{77.64}\scriptsize{$\pm$15.15} \\
S   & 79.91\scriptsize{$\pm$16.48}$^*$ & 79.51\scriptsize{$\pm$23.43}$^*$ & 79.46\scriptsize{$\pm$17.63}$^*$ & 80.18\scriptsize{$\pm$22.13}$^*$ & {81.82}\scriptsize{$\pm$22.22} & \textbf{81.84}\scriptsize{$\pm$22.43} \\ \hline
\end{tabular}
\end{adjustbox}
\caption{Experimental results on the \textit{test} set in  mDSC (\%) with bi- and tri-modal clients.
The local models are trained locally on the private data of each site.
Avg indicates the clients' average performance and ``S'' indicates server performance.
*: $p < 0.05$ comparing against our method.}
\label{tab:multimodal}
\end{table*}

\begin{table*}[!t]
\centering
\begin{adjustbox}{width=1.4\columnwidth}
\begin{tabular}{ccccccc}
\hline
    & Loc. models                     & FedAvg                           & CreamFL                          & FedMSplit                        & FedIoT                           & Ours                             \\ \hline
Avg & 60.54\scriptsize{$\pm$14.89}$^*$ & 56.47\scriptsize{$\pm$15.03}$^*$ & 63.48\scriptsize{$\pm$15.74}$^*$ & 64.72\scriptsize{$\pm$15.81}$^*$ & 64.26\scriptsize{$\pm$15.12}$^*$ & \textbf{67.26}\scriptsize{$\pm$15.34} \\
S   & 79.91\scriptsize{$\pm$16.48}$^*$ & 77.72\scriptsize{$\pm$24.98}$^*$ & 79.95\scriptsize{$\pm$23.14}$^*$ & 80.42\scriptsize{$\pm$22.08}$^*$ & 81.37\scriptsize{$\pm$22.66} & \textbf{81.98}\scriptsize{$\pm$22.68} \\ \hline
\end{tabular}
\end{adjustbox}
\caption{Experimental results on the \textit{test} set in mDSC (\%) with more heterogeneous clients: two clients have one modality (FLAIR and T1c), and the other two have two (T1\&T2 and FLAIR\&T1), with no patient overlap.
The local models are trained locally on the private data of each site.
Avg indicates the clients' average performance and ``S'' indicates server performance.
*: $p < 0.05$ comparing against our method.}
\label{tab:hetero}
\end{table*}

\begin{table}[h]
\centering
\begin{adjustbox}{width=1.\columnwidth}
\begin{tabular}{ccccccc}
\hline
    & Local models                     & FedAvg                           & CreamFL                          & FedMSplit                        & FedIoT                           & Ours                             \\ \hline
Avg & 48.72\scriptsize{$\pm$15.77}$^*$ & 41.07\scriptsize{$\pm$12.76}$^*$ & 38.03\scriptsize{$\pm$17.71}$^*$ & 46.27\scriptsize{$\pm$12.89}$^*$ & 51.68\scriptsize{$\pm$15.13} & \textbf{52.94}\scriptsize{$\pm$14.43} \\
S   & 79.18\scriptsize{$\pm$15.45}$^*$ & 72.78\scriptsize{$\pm$20.02}$^*$ & 79.40\scriptsize{$\pm$15.38}$^*$ & 78.79\scriptsize{$\pm$16.22}$^*$ & 80.56\scriptsize{$\pm$14.87} & \textbf{80.71}\scriptsize{$\pm$15.02} \\ \hline
\end{tabular}
\end{adjustbox}
\caption{Experimental results on {BraTS 2018  \cite{bakas2018identifying}} in mDSC (\%) following setting 1 in the main text.
The local models are trained locally on the private data of each site.
Avg indicates the clients' average performance and ``S'' indicates server performance.
*: $p < 0.05$ comparing against our method.}
\label{tab:brats18}
\end{table}

\section{More Results}
\subsubsection{Subregional Performance:}
In Table \ref{tab:subregion}, we provide class-wise results of different tumor subregions for the setting-1 results shown in Table 2.

\subsubsection{Performance with Eight Monomodal Clients:}
We conduct a more challenging experiment with eight clients, two for each of the four modalities, i.e., $N_m=2$ for $m\in$ \{T1, T1c, T2, FLAIR\}.
Concretely, we keep the server's training data the same as setting 1 in Table 2 and evenly distribute the rest training data to the clients.
The results are shown in Table \ref{tab:8_clients}.
Compared to Table 2, both the clients' and server's performances drop due to the more challenging setting, as expected.
Despite that, the comparative trends are similar.
Notably, our method performs best in both the clients' average and the server's mDSCs.
In addition, it also yields the highest mDSCs for six of eight clients and the second-highest for the rest two clients.
These results further confirm the efficacy of our method in federated learning for medical image analysis in the presence of inter-modal heterogeneity due to missing modalities.

\subsubsection{Performance with Bi- and Tri-modal Clients:}
Our proposed FedMEMA framework can be readily extended to other application scenarios, including the more common ones of multimodal clients.
The extension is straightforward: for a multimodal client, we only need to employ a modality-specific encoder for each of its modalities and a fusion decoder—similar to the server. 
For proof of concept, we experiment with bi- and tri-modal clients here.
Concretely, we keep the server’s training data the same as setting 1 in Table 2 and evenly distribute the rest of the training data to the clients.
Given four modalities, there can be six and four different combinations of two and three modalities, respectively.
Therefore, we use six and four clients in the bi- and tri-modal settings, respectively.
The results in Table \ref{tab:multimodal} show that our method is consistently effective for bi- and tri-modal clients.

\subsubsection{Performance with More Heterogeneous Multimodal Scenario:}
Next, we experiment with a more heterogeneous multimodal scenario where the number of modalities varies for clients: two clients have one modality (FLAIR and T1c), and the other two have two (T1\&T2 and FLAIR\&T1), with no patient overlap.
The results in Table \ref{tab:hetero} demonstrate the advantages of our method over others in the more heterogeneous scenario.

\subsubsection{Performance on Alternative Dataset:}
Following the literature (e.g., Ding et al. 2021),
%'s convention in multimodal brain tumor segmentation 
we use {BraTS 2018 \cite{bakas2018identifying}} as an alternative dataset and present the additional results (corresponding to setting 1 in Table 2) in Table \ref{tab:brats18}.
We can see that the superior performance of our method persists.

\subsubsection{Client-wise Cross-Attention:}
\begin{figure}
\centering
\includegraphics[width=\columnwidth]{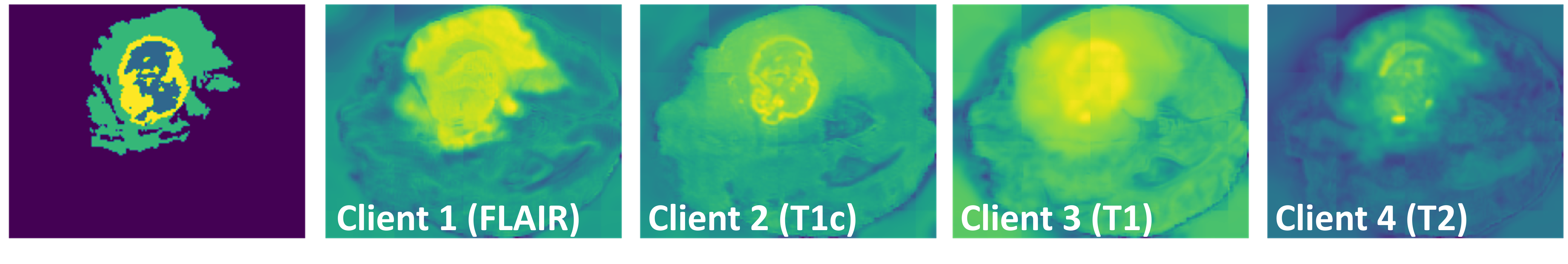}\\
\caption{Client-wise cross-attention, i.e., $F_lA^T_l$ in Eq. (2).
The first column shows the subregional tumor mask.
}
\label{fig:cross-att}
\end{figure}

In the ablation study, rows (d) and (e) in Table 3 demonstrated that our proposed localized adaptive calibration via cross-attention (LACCA) brought substantial performance improvements.
Here, we further visualize the client-wise cross-attention (i.e., $F_lA^T_l$ in Eq. (2)) in Fig. \ref{fig:cross-att},
where the four clients attend to distinct tumor regions in accordance with their modalities.

%\subsubsection{Client-wise Cross-Attention:}
%In the ablation study, rows (d) and (e) in Table 3 demonstrated that our proposed localized adaptive calibration via cross-attention (LACCA) brought substantial performance improvements.
%Here, we further visualize the client-wise cross-attention (i.e., $F_lA^T_l$ in Eq. (2)) in Fig. \ref{fig:cross-att},
%where the four clients attend to distinct tumor regions in accordance with their modalities.
% the tumor region, where the Client 1 (FLAIR) and Client 4 (T2) highlight peritumoral edema and Client 2 (T1c) and Client 4 (T1) highlight Tumor core.

% the tumor region, where the Client 1 (FLAIR) and Client 4 (T2) highlight peritumoral edema and Client 2 (T1c) and Client 4 (T1) highlight Tumor core.

\end{document}